\pdfoutput=1
\documentclass[11pt, a4paper, google]{google}

\usepackage[authoryear, sort&compress, round]{natbib}
\usepackage{longtable}
\usepackage{placeins,afterpage,wrapfig}
\usepackage{xurl}

\newcommand{\system}{ReproBridge}
\newcommand{\bench}{ReproBridge-Bench}
\definecolor{tablehead}{HTML}{E8EEF2}
\definecolor{tablerefined}{HTML}{EAF3EE}
\definecolor{tableink}{HTML}{285D50}
\newcommand{\best}[1]{\textbf{\textcolor{tableink}{#1}}}
\newcolumntype{Y}{>{\raggedright\arraybackslash}X}


\titlespacing*{\paragraph}{0pt}{1.6ex plus .5ex minus .2ex}{1em}
\hypersetup{colorlinks=true,linkcolor=blue!50!black,citecolor=blue!50!black,urlcolor=blue!50!black,
 pdftitle={ReproBridge: Building Reproducibility into ML Research Artifacts},
 pdfauthor={Yubo Wang, Rui Meng, Jaehyun Nam, Tomas Pfister, Jinsung Yoon}}

\uselogo{}  

\title{ReproBridge: Building Reproducibility into ML Research Artifacts}

\author[1,2]{Yubo Wang}
\author[2]{Rui Meng}
\author[2]{Jaehyun Nam}
\author[2]{Tomas Pfister}
\author[2]{Jinsung Yoon}
\affil[1]{University of Waterloo}
\affil[2]{Google}

\begin{abstract}
Reproducing machine learning experiments requires reconstructing procedures from information scattered across papers, released code, and supporting artifacts. For coding agents, reproduction success can therefore depend on their ability to fill in missing details, rather than on the released artifacts alone. We ask whether preserving recovered procedures and decisions improves subsequent independent reproduction. We define a standardized artifact design comprising a refined codebase, environment setup, experiment launchers, and a reproduction guide, and introduce \system{}, an agent workflow for constructing these refined reproducible artifacts. \bench{} provides 1,029 tasks from 387 experiments across 36 papers, anchored by five human-verified reference scores per task. Agents attempt tasks in fresh sessions with raw code or refined artifacts; a separate Auditor checks execution evidence. Across four evaluated models, refined artifacts raise mean reproduction success from 76.4\% to 92.6\% and reduce downstream API spending by 19.7\% on average. The cross-model success-rate spread narrows from 21.6 to 3.9 percentage points, with all four models achieving mean success rates above 90\%. Removing any component reduces success in the ablation study, with the largest losses from code and environment removal. These results suggest that preserving execution knowledge helps even models with lower raw-code success reproduce experiments more reliably and at lower API cost.
\unskip\footnotemark
\end{abstract}

\begin{document}
\maketitle
\footnotetext{Project page: {\urlstyle{same}\url{https://wyyyb.github.io/reprobridge_website/}}}

\afterpage{%
\begin{figure}[!t]
\centering
\includegraphics[width=0.96\linewidth]{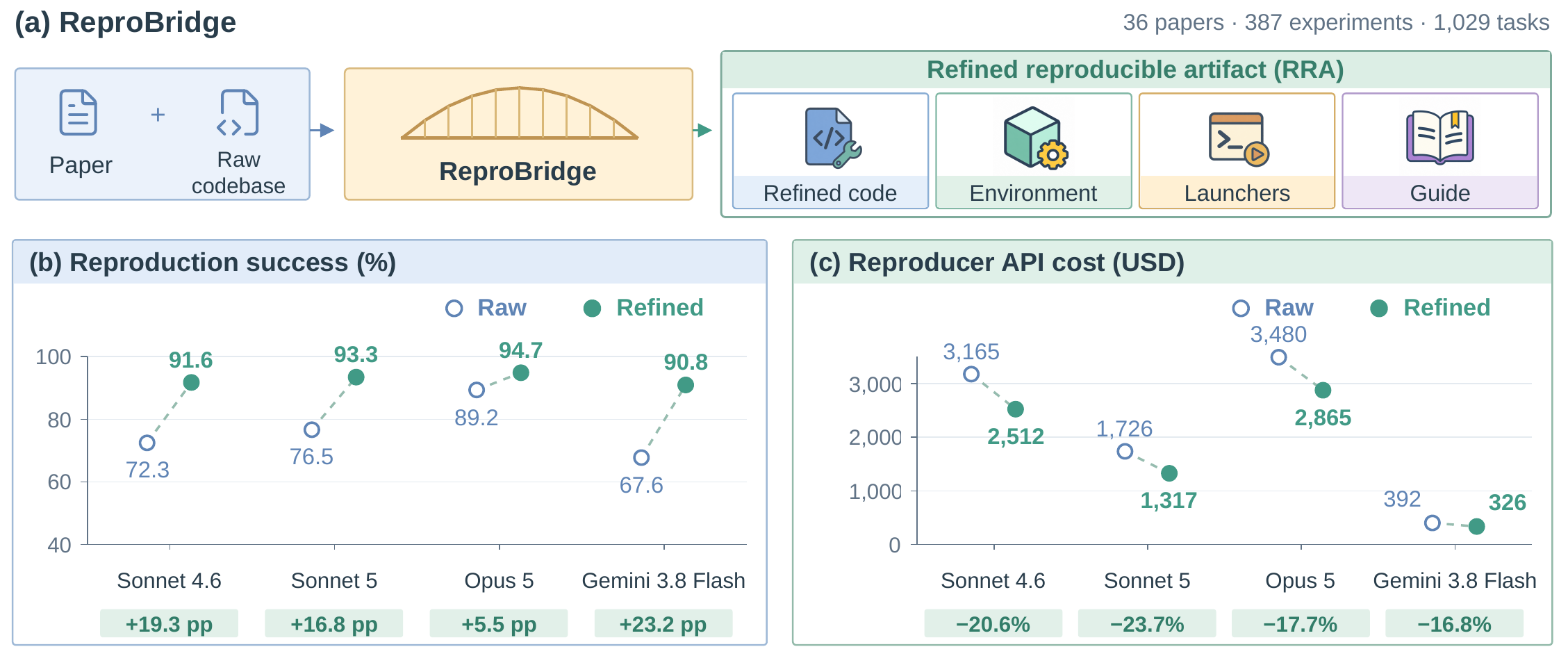}
\caption{\textbf{Preserving reproduction work for reuse.} (a)~\system{} turns a paper and raw codebase into a refined reproducible artifact (RRA) with code, environment setup, launchers, and a guide. (b,c)~Hollow blue (Raw) and filled green (Refined) circles show mean success and Reproducer API spending over three evaluations; dashed lines connect the two conditions for each model. Gain is the success difference (percentage points); cost reduction is relative to Raw. These procedures improve success and reduce reconstruction effort.}
\label{fig:teaser}
\end{figure}}

\section{Introduction}
Reproducing a machine learning experiment often requires recovering details that its paper and released code leave implicit. A reproducer may resolve incompatible dependencies, identify the intended configuration, repair a data loader, and work out how run outputs become a reported score. Even a successful attempt can leave these decisions buried in conversations or shell histories, forcing the next reproducer to repeat the investigation. Code release is an essential starting point \citep{pineau2021reproducibility}; preserving the procedure recovered during execution is a further step toward reusable research artifacts \citep{stodden2016enhancing}.

Coding-agent benchmarks expose this reconstruction burden through unresolved dependencies and misidentified outputs \citep{siegel2024corebench}, missing implementation details \citep{xiang2025scireplicate}, and plans that never become completed experiments \citep{starace2025paperbench}. Consulting cited papers can recover some missing details \citep{zhao2026autoreproduce}, but successful execution may still depend on knowledge a capable agent supplies implicitly. \emph{How much does reproduction success depend on released artifacts, and how much on the reproducer's ability to compensate for missing details?}\looseness=-1

Here, \emph{reproduction} means independently re-executing machine-learning experiments using author-released materials, targeting numerically compatible measurements. \emph{Research artifacts} comprise the code, data, checkpoints, environment specifications, and documentation supporting a study. \textbf{Our hypothesis is that preserving the procedures and decisions recovered during reproduction can improve computational reproducibility.} Dependency resolution can become an installation procedure; experiment discovery, a dedicated launcher; and a correction, a documented choice with its justification and affected tasks. We study this hypothesis through three contributions.\looseness=-1

\paragraph{Artifact design.}
We standardize a \emph{refined reproducible artifact} with four components: a refined codebase, environment setup, experiment launchers, and a reproduction guide (Figure~\ref{fig:teaser}; Section~\ref{sec:artifact_design}). Together they preserve executable procedures and the reasons for their settings and repairs. Component removals test the contribution of each part, while an ordinary-outputs control tests reuse of code, scripts, and logs without the standardized handoff.

\paragraph{Construction workflow.}
\system{} constructs these artifacts from a paper and its official repository (Section~\ref{sec:construction}). Humans define the workflow, delivery standards, and protocols; agents specify experiments, prepare and execute them, and preserve the resulting procedures through review and packaging. Subsequent reuse is evaluated in fresh Reproducer sessions that receive either the raw codebase or the refined artifact.

\paragraph{Benchmark and empirical evaluation.}
\bench{} provides 1,029 tasks from 387 experiments across 36 papers, with five human-verified reference scores per task (Section~\ref{sec:benchmark}). Numerical compatibility is assessed against observed executions and their variability; a separate Auditor checks execution evidence because matching numbers alone can conceal substituted measurements. Across four coding agents, refined artifacts improve success by 5.5 to 23.2~pp and reduce downstream API spending by 16.8\% to 23.7\%. Robustness analyses, component ablations, case studies, and human validation of the integrity checks examine the scope and mechanisms of these gains (Section~\ref{sec:experiments}).

\FloatBarrier
\section{Problem formulation and \bench{}}
\label{sec:benchmark}

\subsection{Execution deviations and reference variation}
\label{sec:error_model}
A task specifies the data, model configuration, training or evaluation procedure, and reporting statistic to reproduce. The paper's stated setup takes precedence over conflicting code. Let $\mu$ be the expected value of that statistic when the specified experiment is faithfully executed. The published score is one observation, not an error-free value of $\mu$.

For an attempt producing a measurement $Y$, we write
\[
Y=\mu+D+\varepsilon.
\]
Here $D$ is the shift in expectation from execution choices departing from the specified experiment, and $\varepsilon$ is zero-mean variation conditional on them. This working model draws on the distinction between systematic and random effects in measurement \citep{jcgm2020models,taylor1994uncertainty}. Stochastic training can vary between runs \citep{nagarajan2018deterministic}. Independent repetitions can reduce uncertainty about the mean of a fixed procedure \citep{bouthillier2021variance}, but do not remove a persistent shift $D$. \system{} aims to reduce $|D|$ by preserving procedures and decisions for faithful execution, while $\varepsilon$ remains.

\label{sec:reference_rationale}
We use five human-verified reference observations of the same statistic per task. Assuming faithful and independent reference executions ($D=0$), their mean estimates $\mu$, and their spread characterizes $\varepsilon$. A prediction interval accommodates uncertainty in that estimated center and variation in a new observation $Y$ \citep{nistprediction}. Section~\ref{sec:references} describes collection and verification. Numerical compatibility is combined with the execution-integrity checks in Section~\ref{sec:audit}: a nearby number alone does not show that the intended experiment was performed.

\subsection{Selection and units of analysis}
\begin{figure}[!t]
\centering
\includegraphics[width=\linewidth]{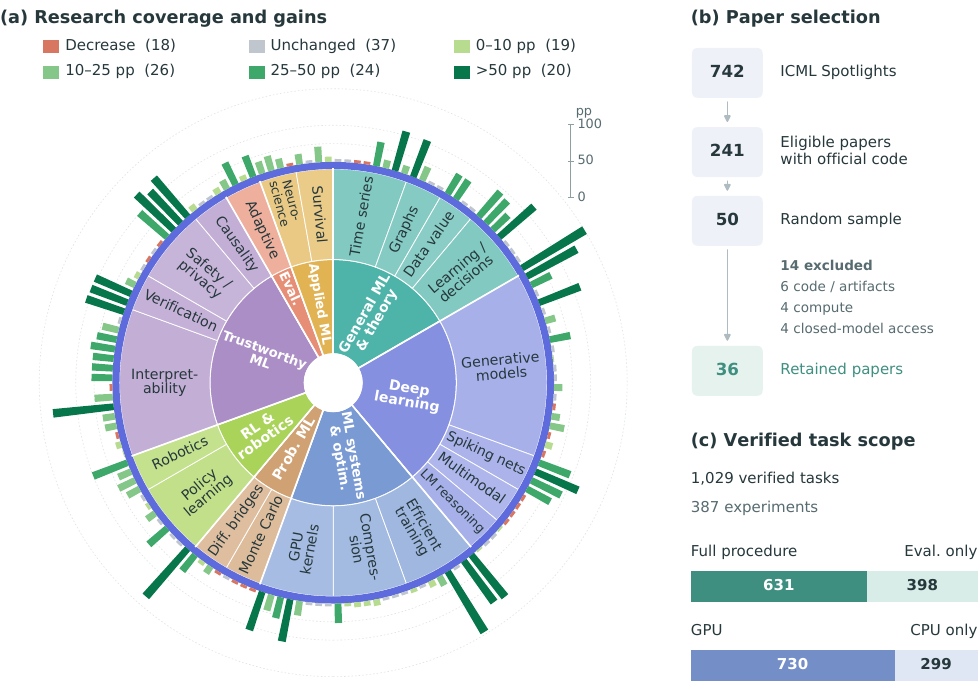}
\caption{\textbf{Research coverage and benchmark composition.} (a)~Eight ICML-adapted topic families and 22 subdomains cover 36 papers; sector angles encode paper counts. Outer marks show 144 paper--model pairs from the first evaluation with Codex auditing. Each paper's four clockwise marks follow Sonnet~4.6, Sonnet~5, Opus~5, and Gemini~3.8 Flash. Green bars encode RRA-minus-Raw gains from the blue zero ring; dashed rings mark 25, 50, and 100~pp. Red/grey ticks indicate decreases/ties without magnitude. Legend counts are pairs; positive bins are $(a,b]$. (b)~Paper selection. (c)~Procedure and hardware are separate partitions of the same 1,029 verified tasks.}
\label{fig:domains}
\end{figure}
The collection targets main experimental results with accessible official implementations. The sampling manifest starts with 742 ICML 2026 Spotlight papers, identifies 241 with code and non-purely-theoretical content, and randomly samples 50. Excluding six for artifact or code availability, four for compute requirements, and four for closed-model access yields 36 retained test papers. Figure~\ref{fig:domains} summarizes their research areas and paired reproduction gains. ReproBridge and evaluation workflows were developed on ten papers, five from NeurIPS~2026 and five from ICLR~2026, scored under the same task, reference, and audit conventions and reported separately in Appendix~\ref{app:development_results}.\looseness=-1

A \emph{task} is one own-method measurement in a main table, identified by dataset, metric, reporting statistic, and location; selected main-table variants are retained. Baselines, stand-alone ablations, figures, and appendix-only results are excluded. An \emph{experiment} can emit several task scores. The 36 papers yield 1,463 curated tasks from 502 experiments. Table~\ref{tab:categories} reports the full verification distribution: 1,029 tasks (70.3\%) have numerically compatible reference results and form the downstream evaluation set, spanning 387 experiments. The remaining 434 tasks distinguish numerical disagreement from execution and resource barriers.

\subsection{Reference collection and task verification}
\label{sec:references}
The reference protocol specifies five human reproducers on separate servers, starting from the same pinned raw commit with separate execution directories and coding-agent sessions. They must not reuse another participant's generated scripts or measurements, refined artifacts, or downstream reproduction traces. Coding-agent assistance is permitted, but humans inspect the scripts, configurations, edits, and measurement extraction actually used. Before aggregation, the five reproducers' execution setups were cross-verified against the paper's specified setup and the common task definition. Each reference observation measures the specified task statistic: a multi-seed mean or best-of-$N$ statistic is one observation, not interchangeable with single-seed scores.\looseness=-1

Verifiers must fix unspecified repetition and seed choices before inspecting scores, and must not tune procedures or repeat valid measurements until they approach the published value. Diagnostic runs and corrected reruns are recorded separately with reasons for replacement; numerical disagreement alone does not justify discarding a valid execution. Appendix~\ref{app:human_reference_protocol} gives the complete verification and evidence-retention protocol.

\paragraph{Numerical compatibility.}
\label{sec:reference_variation}
Following Section~\ref{sec:reference_rationale}, the five observations estimate $\mu$ and reference variation $\varepsilon$. For task $j$, let $p_j$ be the published score and $x_{j1},\ldots,x_{j5}$ the reference observations, with sample mean $\bar{x}_j$ and sample standard deviation $s_j$. A reported range is represented by its midpoint. The compatibility interval is
\begin{equation}
 I_j^{(q)}=[\bar{x}_j-r_j^{(q)},\bar{x}_j+r_j^{(q)}],\qquad
 r_j^{(q)}=\max\left\{t_{(1+q)/2,4}s_j\sqrt{1+\tfrac15},\;0.01|\bar{x}_j|\right\}.
 \label{eq:interval}
\end{equation}
The default is $q=0.95$. Without the explicit 1\% tolerance floor, this is a Student-$t$ prediction interval for a new observation of the same statistic under an independent, approximately normal working model \citep{nistprediction}. It accommodates both uncertainty in the reference mean and variation in the new observation. Thresholds are fixed across raw and refined-artifact conditions; $q=0.80$ supplies the stricter scoring check in Section~\ref{sec:robustness}.

\paragraph{Reference variability and task eligibility.}
Categories 1 and 2 both require $p_j\in I_j^{(.95)}$, and differ only in relative reference variability. We define
\begin{equation}
 V_j=100\,\frac{\max_i x_{ji}-\min_i x_{ji}}{|p_j|}\quad(p_j\ne0);
 \qquad V_j=+\infty\quad(p_j=0).
 \label{eq:variation}
\end{equation}
Stable tasks satisfy $V_j\le10$, meaning that the reference range is at most 10\% of the absolute published score; higher-variability tasks exceed this threshold. Both categories enter the downstream evaluation. We pool them for the main comparison and examine them separately in Section~\ref{sec:variability}. Appendix~\ref{app:reference_rules} gives the uncertainty derivation and additional details of the numerical rules.

\begin{table}[!t]
\centering\small
\renewcommand{\arraystretch}{1.12}
\caption{\textbf{Reference verification outcomes.} A task is one selected own-method measurement from a paper's main tables. Counts and shares cover all 1,463 curated tasks from 36 papers. Categories 1 and 2 have verified reference measurements compatible with the published score and differ at the 10\% relative-variation threshold; together they supply the 1,029 downstream tasks. Categories 3--5 distinguish numerical disagreement, execution blockers, and resource/access limits.}
\label{tab:categories}
\begin{tabularx}{\linewidth}{clYrr}
\toprule
\rowcolor{tablehead}
\textbf{Cat.} & \textbf{Verification outcome} & \textbf{Interpretation} & \textbf{Tasks} & \textbf{Share (\%)} \\
\midrule
1 & Reproduced: stable & Compatible reference, low variability & 701 & 47.9 \\
2 & Reproduced: variable & Compatible reference, higher variability & 328 & 22.4 \\
3 & Numerical mismatch & Executed, but numerically incompatible & 183 & 12.5 \\
4 & Execution blocked & Missing artifacts, dependencies, or code defects & 186 & 12.7 \\
5 & Resource/access limit & Compute capacity or restricted access & 65 & 4.4 \\
\midrule
\rowcolor{tablerefined}
\multicolumn{3}{l}{\textbf{Downstream evaluation (Categories 1 + 2)}} & \best{1,029} & \best{70.3} \\
\multicolumn{3}{l}{Total curated tasks} & 1,463 & 100.0 \\
\bottomrule
\end{tabularx}
\end{table}

\subsection{Scope and information access}
The evaluated set contains 631 full-procedure and 398 evaluation-only tasks (730 GPU, 299 CPU-only). Evaluating an explicitly permitted official checkpoint does not require repeating its training. Evaluation alone cannot establish completion of a training task.

All eight downstream conditions use the same task categories and numerical thresholds, applied outside the Reproducer session. Published targets remain visible in the paper and task specification; the Auditor checks that submitted values come from new executions. Session instructions withhold the reference information, alternative repository condition, and other agents' run directories. An allowed checkpoint may serve as input; a shipped score file never counts as a measurement.

The benchmark includes 586 lower-is-better and 443 higher-is-better measurements, all assessed by the same two-sided compatibility rule. This evaluates agreement with the specified experimental result: a score beyond either bound remains numerically incompatible, including one that improves on the published value. Reference submissions must also link each score to the starting code commit, executed scripts and edits, environment and hardware, configuration and seeds, raw outputs, and score-extraction or aggregation procedure, making the reported statistic traceable to a concrete execution. Appendix~\ref{app:selection} gives selection records, verification protocols, and statistical rules.\looseness=-1

\FloatBarrier
\section{ReproBridge: artifact design and construction}
\label{sec:method}
\system{} turns reproduction work into an artifact that a later researcher or agent can use without access to the construction conversation. The handoff comprises refined code, environment setup, experiment launchers, and a reproduction guide. Its construction workflow preserves the evidence behind execution choices as well as the executable steps themselves.

\subsection{Construction workflow}
\label{sec:construction}
\begin{figure}[!htb]
\centering
\includegraphics[width=\linewidth]{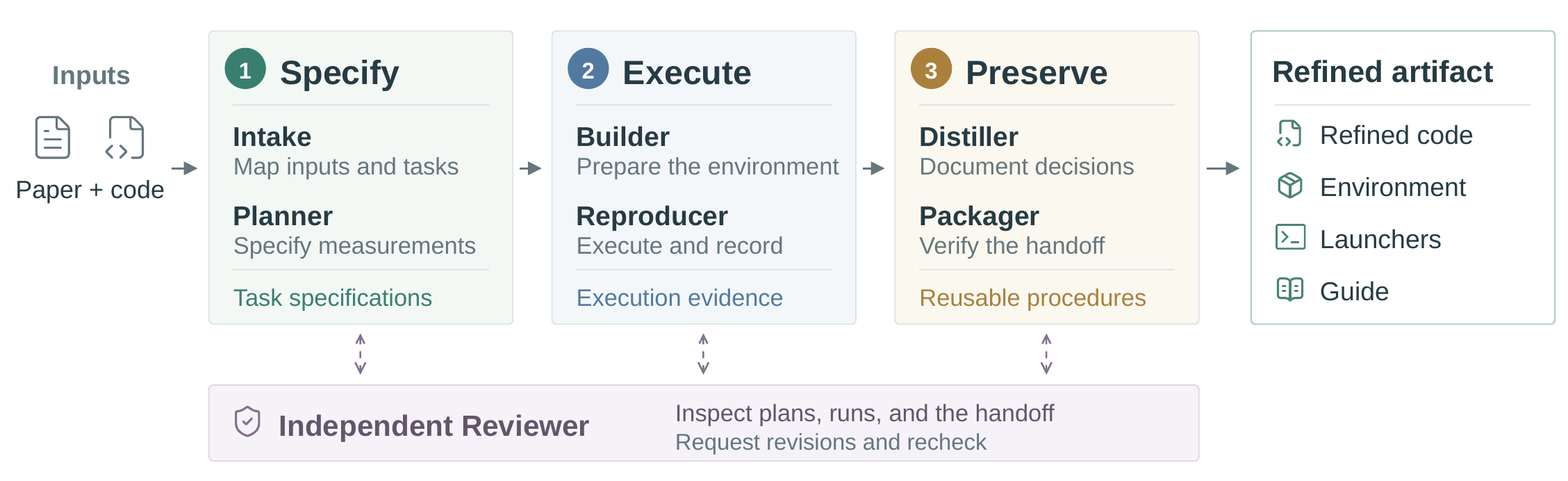}
\caption{\textbf{ReproBridge artifact construction.} Six roles form three stages: Specify (Intake, Planner), Execute (Builder, construction Reproducer), and Preserve (Distiller, Packager). Footers list stage outputs. Solid arrows show construction; dashed links show independent review and revision. The final handoff contains refined code, environment setup, launchers, and a guide. Downstream reproduction and auditing are separate evaluation steps.}
\label{fig:overview}
\end{figure}

Humans define the workflow, delivery standards, and protocols; agents perform all construction, review, and packaging. Figure~\ref{fig:overview} organizes this work into three stages: \emph{Specify} fixes the intended experiment before execution; \emph{Execute} establishes a working procedure and records measurements, repairs, and failures; and \emph{Preserve} packages that procedure and its rationale for reuse. Table~\ref{tab:construction_roles} specifies each role's inputs and persistent outputs.

\begin{table}[htbp]
\centering\small
\caption{Construction skills: inputs, responsibility, and persistent outputs. A checkpoint reviewer is reused at different stages in separate contexts. The construction Reproducer is not the fresh downstream session used to evaluate reuse.}
\label{tab:construction_roles}
\begin{tabularx}{\linewidth}{@{}lYY@{}}
\toprule
\rowcolor{tablehead}
Role & Input and responsibility & Output and check \\
\midrule
Intake & Paper, pinned raw code, selected tasks; read and map the inputs & Input manifest, experiment groups, artifact availability, initial ambiguities \\
Planner & Task mapping and ambiguities; specify the measurement before execution & Task specifications with source references, recorded choices, execution plans, draft launchers \\
Builder & Dependency and artifact requirements; construct a working environment & Locked dependencies, setup and data scripts, container definition, smoke-test evidence \\
Reproducer & Plans and verified setup; execute and record justified changes & File-backed measurements, code changes, repetition records, explicit failure reasons \\
Distiller & Execution evidence, repairs, and decisions; synthesize the handoff & Reproduction guide, change record, aggregation and variability guidance \\
Packager & Code, setup, launchers, and documents; assemble a portable deliverable & Refined artifact and clean-shell verification evidence \\
Reviewer & Plans, execution records, or completed artifact; challenge unsupported claims & Localized findings at planning, reproduction, and final handoff checkpoints \\
\bottomrule
\end{tabularx}
\end{table}

\paragraph{Persistent evidence connects the stages.}
A phase controller sequences the roles and checks required outputs between stages. Before scored execution, task specifications record the data, model, procedure, reporting statistic, source references, and unresolved choices. This makes execution decisions explicit before their results are known. When paper and code conflict, agents follow the paper's setup and flag the discrepancy (Section~\ref{sec:discussion}). Each measurement must link to an actual output file; blocked tasks retain an explicit failure reason, and diagnostic runs remain separate from scored repetitions. The guide and change record are then grounded in these stored specifications, execution records, and justified repairs.

\paragraph{Independent review tests the evidence and the handoff.}
A Reviewer inspects plans, execution records, and the completed artifact in contexts separate from the roles being reviewed. It returns localized findings to the responsible stage for revision and rechecking. Delivery also requires testing the supplied entry points from a clean shell, including an inexpensive real repetition and checks on output structure and score extraction. The delivered environment must match the one used for construction measurements. Human reference information is withheld throughout construction. Appendix~\ref{app:implementation} gives the full role specifications and delivery checks.

\subsection{Four components preserve execution knowledge}
\label{sec:artifact_design}
A later reproducer needs to know what code to run, how to prepare its environment, how to obtain the requested measurement, and why particular choices were made. The four-component \emph{refined reproducible artifact} (RRA in figures and tables) addresses these needs. The \emph{refined codebase} repairs execution defects and supplies missing experiment support. \emph{Environment setup} specifies compatible dependencies and data or checkpoint acquisition. \emph{Experiment launchers} connect tasks to configurations, execution scope, outputs, and measurement extraction, distinguishing scored runs from smoke tests. The \emph{reproduction guide}, including its change record, explains settings, paper--code discrepancies, reporting statistics, and resource or numerical sensitivities. Section~\ref{sec:ablation} tests the contribution of each component through removal experiments.

\begin{figure}[!htb]
\centering
\includegraphics[width=\linewidth]{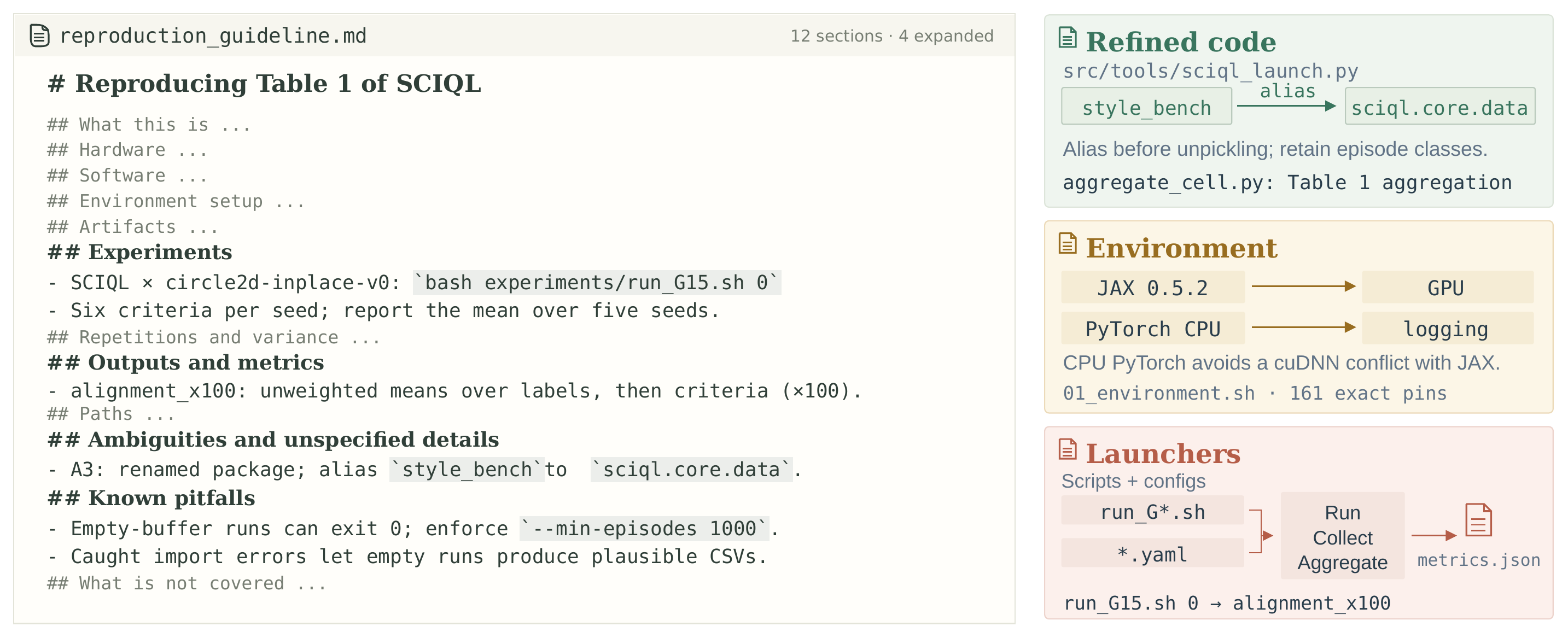}
\caption{\textbf{A refined artifact for SCIQL.} Left: selected guide entries specify the six-criterion, five-seed protocol, unweighted aggregation, and checks for silently empty datasets. Right: an alias applied before unpickling preserves episode classes; CPU PyTorch avoids a cuDNN conflict with GPU JAX; launchers connect configurations to execution, output collection, and aggregation. The four components preserve executable steps and the reasons for them.}
\label{fig:guide_main}
\end{figure}

The guide connects instructions to their rationale. Each ambiguity entry records its source location, the gap, the adopted resolution and rationale, sensitivity, and affected tasks. The change record distinguishes execution fixes, correctness fixes, and convenience changes, while retaining pitfalls and open choices. Repetition guidance preserves the specified reporting unit and characterizes variation $\varepsilon$ without selecting favorable outcomes. Figure~\ref{fig:guide_main} illustrates how SCIQL's guide links measurement decisions and failure checks to executable support; Appendix~\ref{app:guide_example} expands its twelve-section structure.

\section{Experiments}
\label{sec:experiments}
\subsection{Evaluation protocol and setup}
\label{sec:audit}
Opus~5 constructs the refined artifacts for the main experiment. Fresh Reproducer sessions receive the paper, selected tasks, and either raw code or its refined artifact. Construction conversations, hidden reference observations, and the alternative condition are withheld. Sessions submit new measurements and execution evidence; we compare success, token usage, and API spending.

Sonnet~4.6, Sonnet~5, Opus~5, and Gemini~3.8 Flash (Gemini~3.8 in figures and tables) attempt all 1,029 tasks from 36 test papers, with one fresh session per paper--model--condition per evaluation. Both raw-code and refined-artifact conditions use a 24-hour wall-clock limit per paper. GPU experiments use eight H100-80GB GPUs. Codex is the primary Auditor. Harnesses differ across vendors, so cross-model comparisons are descriptive. Paper-wise, stratified, and task-level diagnostics use the first evaluation. Appendix~\ref{app:evaluation} details the harness, information boundaries, and auditing.\looseness=-1

\paragraph{Execution-integrity audit.}
An independent Auditor checks the evidence chain for each reported result, tracing the value to its output location, execution record, and generating script. It uses ledger timestamps to check execution and extraction order, and available script and output hashes to verify that the inspected files match the recorded versions. The Auditor cross-checks executed code, logs, and extracted values to locate copied measurements, altered outputs, or substituted computations.\looseness=-1

After mechanical screening identifies leads, a \emph{Finder} inspects executed code and records and proposes localized violations. A separate \emph{Refuter} tries to overturn each proposed invalidation by checking the evidence, its effect on the measurement, and the task specification. Unsubstantiated findings remain uncertain. If the evidence remains contested, a panel examines its support, the measurement effect, and alternative explanations. The Auditor operates after reproduction and provides no feedback to the Reproducer.

\paragraph{Invalidation criteria and task success.}
The five eligible violation classes distinguish copying a value without measuring it, substituting the measurement instrument, altering an operand used by the intended formula, choosing the reported outcome conditional on observed values, and computing the wrong statistic. Invalidation requires an eligible violation supported by localized evidence, a material effect on the measurement, and a failed refutation. A numerical mismatch, crashed run, hardware difference, or suspicion without a localized artifact does not automatically meet these conditions. Paper--code discrepancies are assessed against the same criteria.

A task succeeds when its result lies in the fixed compatibility interval (Equation~\ref{eq:interval}) and is not invalidated by the primary Auditor; missing or incompatible results fail. Uncertain audit findings do not count as invalidations. Section~\ref{sec:audit_results} checks the sensitivity of this scoring rule to the Auditor and validates audit judgments against human review.

\subsection{Refined artifacts improve success and reduce downstream cost}
\label{sec:primary}
\label{sec:cost_results}
Refined artifacts improve success and lower downstream cost for every tested model (Table~\ref{tab:main}). Gains range from 5.5~pp for Opus~5, the strongest raw reproducer, to 23.2~pp for Gemini, with 6.3\% to 19.8\% fewer tokens and 16.8\% to 23.7\% lower API spending, including failed attempts. Cross-model summaries weight models equally; API savings are relative to each model's raw-code cost.\looseness=-1

\begin{table}[!htb]
\centering\small
\setlength{\tabcolsep}{5.7pt}
\renewcommand{\arraystretch}{1.25}
\caption{\textbf{Main results on 36 test papers and 1,029 tasks.} Raw uses the released codebase; RRA uses the refined reproducible artifact. Entries report mean $\pm$ standard deviation over three evaluations (SD in subscripts). Success requires numerical compatibility and no confirmed audit violation. Gain is RRA minus Raw in percentage points; Red. is $100(\mathrm{Raw}-\mathrm{RRA})/\mathrm{Raw}$, computed from the means. Tokens (B: billions) and API USD sum Reproducer sessions, including failed attempts; construction, auditing, \mbox{and GPU costs are excluded.}\looseness=-1}
\label{tab:main}
\begin{tabular}{@{}l r >{\columncolor{tablerefined}}r r r >{\columncolor{tablerefined}}r r r >{\columncolor{tablerefined}}r r@{}}
\toprule
\rowcolor{tablehead}
 & \multicolumn{3}{c}{\textbf{Success (\%)}} & \multicolumn{3}{c}{\textbf{Tokens (B)}} & \multicolumn{3}{c}{\textbf{API (USD)}} \\
\cmidrule(lr){2-4}\cmidrule(lr){5-7}\cmidrule(l){8-10}
\textbf{Model} & Raw & \textbf{RRA} & Gain & Raw & \textbf{RRA} & Red. & Raw & \textbf{RRA} & Red. \\
\midrule
Sonnet 4.6 & $72.3_{\pm 1.1}$ & $\mathbf{91.6}_{\pm 3.4}$ & +19.3 & $8.48_{\pm 0.38}$ & $6.80_{\pm 0.76}$ & 19.8 & $3{,}165_{\pm 142}$ & $2{,}512_{\pm 286}$ & 20.6 \\
Sonnet 5 & $76.5_{\pm 2.8}$ & $\mathbf{93.3}_{\pm 1.4}$ & +16.8 & $4.98_{\pm 0.55}$ & $4.52_{\pm 0.24}$ & 9.2 & $1{,}726_{\pm 175}$ & $1{,}317_{\pm 102}$ & 23.7 \\
Opus 5 & $89.2_{\pm 1.2}$ & $\mathbf{94.7}_{\pm 2.2}$ & +5.5 & $3.01_{\pm 0.16}$ & $2.82_{\pm 0.29}$ & 6.3 & $3{,}480_{\pm 165}$ & $2{,}865_{\pm 312}$ & 17.7 \\
Gemini 3.8 & $67.6_{\pm 1.9}$ & $\mathbf{90.8}_{\pm 1.1}$ & +23.2 & $2.24_{\pm 0.26}$ & $1.83_{\pm 0.09}$ & 18.3 & $392_{\pm 41}$ & $326_{\pm 16}$ & 16.8 \\
 
\bottomrule
\end{tabular}
\end{table}

The cross-model success-rate spread narrows from 21.6 to 3.9~pp. \textbf{This suggests that, given sufficient execution information, Reproducers with lower raw-code success can achieve high success.} This supports our hypothesis that preserving execution knowledge reduces the need to infer missing details. Section~\ref{sec:robustness} examines paper-level heterogeneity and scoring sensitivity.

\paragraph{External benchmark.}
We compare handoffs on 45 CORE-Bench Hard capsules \citep{siegel2024corebench}; Sonnet~5 constructs the refined artifacts and VERITAS's patched codebase and fix log \citep{liu2026veritas} (Table~\ref{tab:corebench}). Refined artifacts achieve the highest success, reduce median duration from 33 to 14 minutes, and lower spending per successful capsule from \$8.43 to \$4.34. VERITAS lies between raw and refined handoffs; the latter matches or exceeds raw success in each area.

\begin{table}[!htb]
\centering\small
\setlength{\tabcolsep}{4pt}
\renewcommand{\arraystretch}{1.12}
\caption{\textbf{Artifact reuse on CORE-Bench Hard.} Sonnet~4.5 reproduces 45 capsules from raw code, VERITAS's patched code and fix log, or a refined artifact (RRA); Sonnet~5 constructs both handoffs. A capsule succeeds only if all its questions satisfy the benchmark's reference intervals (79 questions overall). (a)~Overall capsule success, Reproducer API USD per successful capsule, and median Reproducer session duration. (b)~Success within computer (CS), medical (Med.), and social (Soc.) sciences; parentheses give capsule counts.}
\label{tab:corebench}
\begin{minipage}[t]{0.53\linewidth}
\centering
\textbf{(a) Overall}\\[2pt]
\begin{tabular}{lrrr}
\toprule
\rowcolor{tablehead}
\textbf{Handoff} & Success (\%) & USD/success & Time (min) \\
\midrule
Raw & 62.2 & 8.43 & 33.0 \\
VERITAS & 68.9 & 6.78 & 24.5 \\
\rowcolor{tablerefined}
\textbf{RRA} & \best{73.3} & \best{4.34} & \best{14.4} \\
\bottomrule
\end{tabular}
\end{minipage}\hfill
\begin{minipage}[t]{0.45\linewidth}
\centering
\textbf{(b) Success by area (\%)}\\[2pt]
\begin{tabular}{lrrr}
\toprule
\rowcolor{tablehead}
\textbf{Handoff} & CS (18) & Med. (13) & Soc. (14) \\
\midrule
Raw & 50.0 & 76.9 & 64.3 \\
VERITAS & 61.1 & \best{84.6} & 64.3 \\
\rowcolor{tablerefined}
\textbf{RRA} & \best{66.7} & \best{84.6} & \best{71.4} \\
\bottomrule
\end{tabular}
\end{minipage}
\end{table}

\subsection{Gains persist across papers and scoring thresholds}
\label{sec:robustness}
The main comparison pools task outcomes, so papers with more selected measurements contribute more weight. We examine whether gains persist when papers receive equal weight, when numerical scoring is stricter, and within each reference-variability group. These diagnostics use the first evaluation; Table~\ref{tab:main} reports means over three evaluations.

\begin{figure}[!htb]
\centering
\includegraphics[width=0.98\linewidth]{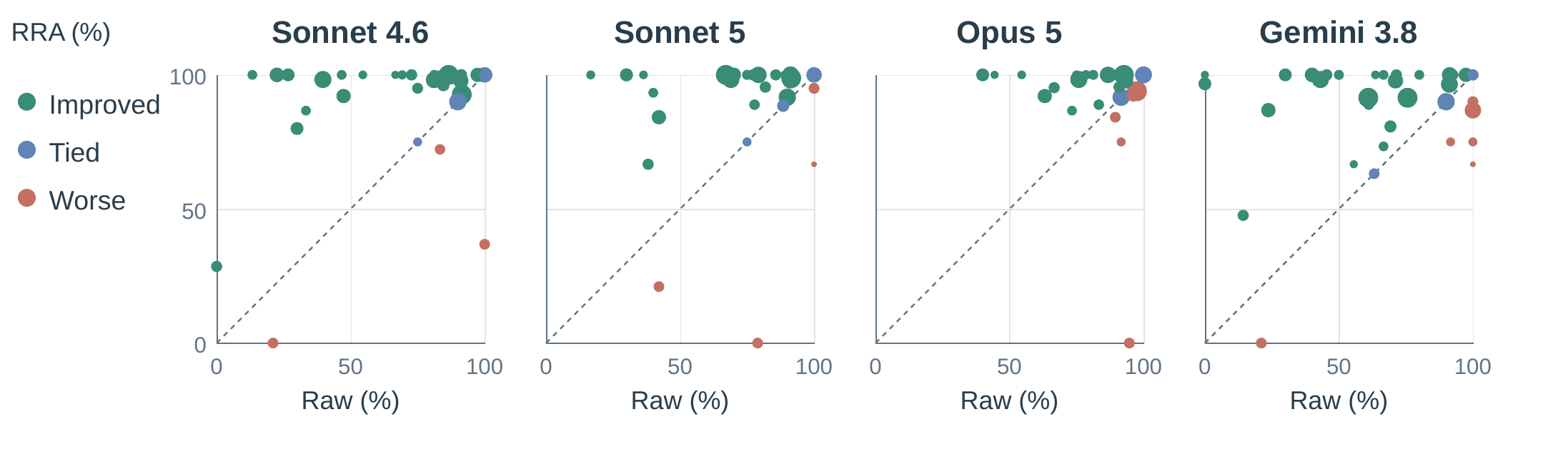}
\caption{\textbf{Paper-wise reproduction success in the first evaluation.} Each panel contains one point per retained paper, comparing the percentage of the same tasks that succeed with Raw (horizontal axis) and the refined artifact, RRA (vertical axis), under primary Codex scoring. Marker area is proportional to task count. Green points above the equality line improve, blue points on the line tie, and red points below it regress. Overlapping observations retain their exact positions.}
\label{fig:paper_scatter}
\end{figure}

\paragraph{Paper-level heterogeneity.}
Figure~\ref{fig:paper_scatter} compares raw and refined-artifact success within each of the 36 papers. Paper-macro rates average these per-paper percentages with equal weight. All four models improve under this aggregation, with gains of 6.7 to 20.7~pp (Table~\ref{tab:robustness}a). Success improves or ties on 30 to 33 papers per model, while three to six papers regress. Thus the aggregate gains extend across papers, although refinement does not improve every paper. Section~\ref{sec:regressions} examines remaining failures and task-level reversals.

\begin{table}[!htb]
\centering\small
\setlength{\tabcolsep}{3pt}
\renewcommand{\arraystretch}{1.18}
\caption{\textbf{Gains persist with equal paper weights and stricter scoring.} First-evaluation results under primary Codex scoring. (a)~Paper-macro success gives equal weight to all 36 papers with the default 95\% interval; I/T/W counts improved, tied, and worse papers. (b)~Task-level success uses the 80\% interval on the same 1,029 tasks, with category labels fixed. Raw and RRA are success percentages; Gain is RRA minus Raw in percentage points, calculated before rounding.}
\label{tab:robustness}
\label{tab:paper_macro}
\label{tab:strict}
\begin{tabularx}{\linewidth}{@{}l*{7}{>{\raggedleft\arraybackslash}X}@{}}
\toprule
 & \multicolumn{4}{c}{\textbf{(a) Paper-macro, 95\% interval}} & \multicolumn{3}{c}{\textbf{(b) Task-level, 80\% interval}} \\
\cmidrule(lr){2-5}\cmidrule(l){6-8}
\rowcolor{tablehead}
Model & Raw & RRA & Gain & I/T/W & Raw & RRA & Gain \\
\midrule
Sonnet 4.6 & 69.9 & 89.7 & +19.8 & 25/8/3 & 61.3 & 88.2 & +26.9 \\
Sonnet 5 & 75.1 & 90.6 & +15.5 & 23/9/4 & 63.9 & 90.5 & +26.5 \\
Opus 5 & 87.5 & 94.2 & +6.7 & 16/15/5 & 82.0 & 92.3 & +10.3 \\
Gemini 3.8 Flash & 67.2 & 87.9 & +20.7 & 25/5/6 & 47.7 & 86.8 & +39.1 \\
 
\bottomrule
\end{tabularx}
\end{table}

\paragraph{Stricter numerical compatibility.}
Replacing the default 95\% interval with the 80\% interval in Equation~\ref{eq:interval} preserves the selected tasks and the primary Auditor's judgments. Refined artifacts still improve success for every model, with gains of 10.3 to 39.1~pp (Table~\ref{tab:robustness}b). The positive gains therefore persist under a tighter numerical criterion.

\paragraph{Reference variability.}
\label{sec:variability}
Both reference-variability groups improve across all four models, with larger point-estimate gains in Cat.~1: 8.0 to 23.3~pp versus 2.1 to 22.3~pp in Cat.~2. The groups differ in task composition and interval width, so this contrast is descriptive. Appendix~\ref{app:additional_results} provides the full stratified results.

\subsection{Preserved procedures support recovery and reduce reconstruction}
\label{sec:mechanisms}
\begin{figure}[!htb]
\centering
\includegraphics[width=0.96\linewidth]{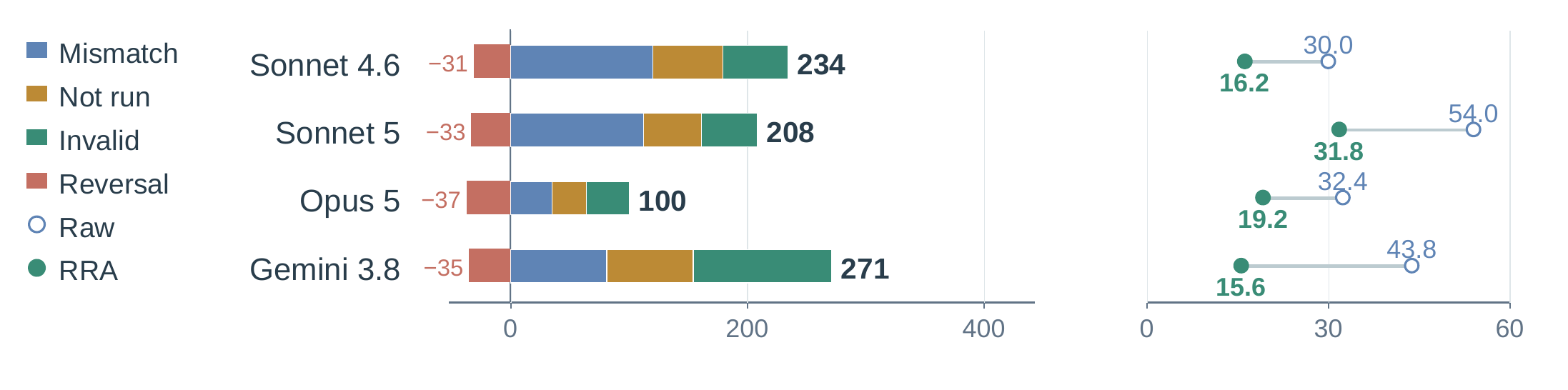}
\caption{\textbf{Recovering failures and shortening preparation.} First-evaluation results across four models and 1,029 tasks per model. Left: positive bars count Raw failures becoming RRA successes, grouped by the Raw outcome: numerical mismatch, execution/resource barrier (Not run), or audit-invalid numerical match. Negative bars count Raw successes lost with RRA; labels give recovery and reversal totals. Right: median time to first scored execution across paper sessions (minutes); hollow and filled markers denote Raw and RRA, respectively.}
\label{fig:transitions}
\end{figure}

Case studies and qualitative analyses use the first evaluation. Across paired task--model outcomes, 813 raw failures become successes, against 136 reversals (Figure~\ref{fig:transitions}). Recoveries include 348 numerical mismatches, 210 unexecuted tasks due to execution or resource barriers (\emph{Not run}), and 255 audit-invalid numerical matches. These records suggest two complementary contributions: completing ambiguous or incomplete measurement specifications and preserving executable procedures for reuse.

\paragraph{Completing and clarifying measurement specifications.}
In Mixture of Concept Bottleneck Experts \citep{santis2026mixture}, the reduction of multiple tree-edit distances is unspecified, and the released code lacks a collector for Table~1. Sonnet~5's raw attempt sums the four distances, reporting 60 rather than 15. The refined artifact resolves this ambiguity by specifying the per-expression mean and supplying the collector; the later attempt reports 15. SCIQL's released code similarly leaves the mapping from run outputs to table entries incomplete; the refined artifact specifies the output field and aggregation across criteria and seeds. Such clarification and experiment-to-result mapping together represent about 35\% of the case-review attribution distribution.\looseness=-1

\paragraph{Recovered configurations become executable: HELIX.}
For HELIX \citep{zhang2026helix}, Opus~5 restores the paper's architecture on raw code but retains the shipped batch size and learning rate. The refined artifact identifies a repository revision matching that architecture, records it as the source of the adopted settings, and exposes them through a launcher. A later attempt produces numerically compatible MAEs for both illustrated tasks. The guide records the evidence for the choice, and the launcher makes it reusable (Appendix~\ref{app:helix_case}).

\paragraph{A compatible environment transfers across sessions: WBMM.}
For six COCO tasks in WBMM \citep{song2026wbmm}, Sonnet~4.6's raw attempt fails to install a compatible vision-library stack for H100. The refined artifact preserves the dependency versions, build prerequisite, and source-build route for the target architecture. The later session executes all six tasks without a code patch; its box and mask AP values agree numerically with the reported targets. This case shows how an environment repair can become a reusable installation procedure (Appendix~\ref{app:wbmm_case}).

\paragraph{Fixing bugs in the released code.}
In Effective Model Pruning \citep{wang2026effective}, the released loader silently labels all 10,000 TinyImageNet validation images as one class. During construction, \system{} adds a script organizing images by their true labels and a launcher check requiring prepared data. In paired Gemini~3.8 Flash runs, ResNet-18 validation loss changes from 8.898 on raw code to 2.288 with the refined artifact, close to the paper's 2.303. Model--paper sessions changing code downstream fall from 113/144 to 54/144, and median time to first scored execution decreases for every model (Figure~\ref{fig:transitions}, right). Together with lower token use, this indicates less reconstruction.\looseness=-1

\paragraph{Explicit procedures also help the model that built them.}
Clearer code organization and explicit execution decisions can reduce repeated search and context demands, even for the model that built the artifact. In ETTFS \citep{che2026efficiently}, the raw Opus~5 attempt searches 6,377 lines of Python for measurement code absent from the release. The refined artifact organizes the added measurement code into a dedicated profiling module, a launcher that sequences training and profiling, and documented output fields. A fresh Opus~5 session follows this structure and completes a previously blocked measurement.

\paragraph{Remaining failures.}
\label{sec:regressions}
Failures shared by both handoffs mainly reflect paper--code setup mismatches, typographical errors in the paper, or experimental variability placing measurements outside the compatibility interval. Refinement can also preserve or introduce \mbox{incorrect experimental choices.}

The 136 reversals from raw success to refined failure comprise 84 numerical mismatches, 31 resource/access failures, 11 audit-invalid matches, and 10 execution blockers. AuxiliaryMCMC and LiME account for 50 and 14 reversals, respectively, or 47.1\% combined. The LiME case \citep{kowsher2026lime} illustrates an attempt that fails to complete training and evaluation; AuxiliaryMCMC \citep{yuan2026markov} illustrates numerical fragility, hardware sensitivity, and protocol deviations under time pressure (Appendix~\ref{app:regressions}).

\Needspace{18\baselineskip}
\subsection{Each artifact component contributes to reuse}
\label{sec:ablation}
\begin{wraptable}{r}{0.53\textwidth}
\vspace{-\intextsep}
\centering\small
\setlength{\tabcolsep}{3pt}
\renewcommand{\arraystretch}{1.15}
\caption{\textbf{Handoff and component ablations.} Sonnet~5, 287 tasks, 12 papers, one evaluation. Loss is the drop from full RRA (pp); removals follow the \mbox{adjacent protocol.}}
\label{tab:ablation}
\begin{tabular}{@{}lrr@{}}
\toprule
\rowcolor{tablehead}
\textbf{Handoff} & Success (\%) & Loss (pp) \\
\midrule
Raw code & 73.9 & --- \\
Ordinary outputs & 81.2 & --- \\
\rowcolor{tablerefined}
\textbf{Full RRA} & \best{93.0} & --- \\
\midrule
\multicolumn{3}{@{}l}{\textit{Full RRA without one component:}} \\
\quad Refined codebase & 87.8 & 5.2 \\
\quad Environment setup & 88.5 & 4.5 \\
\quad Reproduction guide & 89.2 & 3.8 \\
\quad Experiment launchers & 89.9 & 3.1 \\
\bottomrule
\end{tabular}
\vspace{-\intextsep}
\end{wraptable}

Ordinary outputs are Opus~5's raw-code execution directory with code, scripts, and logs retained and result files removed. They carry no standardized guide.

\paragraph{Component-removal protocol.}
Refined-code removal restores the original source; environment removal deletes the setup route; launcher removal deletes the experiment entry points. For these three conditions, descriptions of the removed component are also deleted from the remaining documentation and scripts. The guide ablation removes the guide itself. The change record belongs to the documentation component.\looseness=-1

A source file and an explanation of how to reconstruct it can carry overlapping knowledge. Deleting only the file leaves a route for the downstream agent to restore the same functionality. Removing both the component and its descriptions tests the loss of that information while leaving the other components available; the resulting losses are conditional on those retained materials.

\paragraph{The full artifact improves on ordinary outputs.}
The complete artifact outperforms ordinary outputs by 11.8~pp (Table~\ref{tab:ablation}). Ordinary outputs also outperform raw code, demonstrating value in reusing execution work. Both handoffs come from separate upstream runs with comparable resource use. The comparison evaluates protocol-guided construction and its resulting handoff.

Code and environment removal cause the largest losses, showing the need for repaired implementations and compatible dependencies. Removing the guide or launchers also reduces success when code and environment remain available: an executable codebase can still leave settings, measurement rules, and entry points implicit. Retained components can partly compensate for a removal (Appendix~\ref{app:ablation}). These losses favor pairing executable support with explicit guidance on measuring results.\looseness=-1

\subsection{Gains persist across Auditors despite imperfect detection}
\label{sec:audit_results}
\paragraph{Sensitivity to the Auditor.}
Codex and Claude inspect the same numerically matching outcomes in the first evaluation (Table~\ref{tab:audit_sensitivity}). Refined artifacts improve success under either Auditor for every model. Differences in invalidations change the absolute success rates, while the direction of the raw-to-refined comparison is preserved.

\begin{table}[!htb]
\centering\small
\setlength{\tabcolsep}{4pt}
\renewcommand{\arraystretch}{1.1}
\caption{\textbf{Sensitivity to the Auditor in a single evaluation.} Codex and Claude inspect the same numerically matching outcomes for each model and handoff. Invalid / matched reports the Auditor's invalidations and the number of numerical matches; Success is $100(\mathrm{matched}-\mathrm{invalid})/1{,}029$. Raw denotes the released codebase and RRA the refined artifact.}
\label{tab:audit_sensitivity}
\begin{tabular}{@{}llrrrr@{}}
\toprule
\rowcolor{tablehead}
\textbf{Reproducer} & \textbf{Input} & \multicolumn{2}{c}{\textbf{Codex}} & \multicolumn{2}{c}{\textbf{Claude}} \\
\rowcolor{tablehead}
 & & Invalid / matched & Success (\%) & Invalid / matched & Success (\%) \\
\midrule
Sonnet 4.6 & Raw & 63/805 & 72.1 & 41/805 & 74.2 \\
\rowcolor{tablerefined}Sonnet 4.6 & RRA & 23/968 & 91.8 & 6/968 & 93.5 \\
Sonnet 5 & Raw & 51/838 & 76.5 & 4/838 & 81.0 \\
\rowcolor{tablerefined}Sonnet 5 & RRA & 0/962 & 93.5 & 0/962 & 93.5 \\
Opus 5 & Raw & 43/961 & 89.2 & 0/961 & 93.4 \\
\rowcolor{tablerefined}Opus 5 & RRA & 10/991 & 95.3 & 0/991 & 96.3 \\
Gemini 3.8 & Raw & 173/869 & 67.6 & 119/869 & 72.9 \\
\rowcolor{tablerefined}Gemini 3.8 & RRA & 6/938 & 90.6 & 3/938 & 90.9 \\
 
\bottomrule
\end{tabular}
\end{table}

\paragraph{Validation against human review.}
Two human reviewers cross-verified all 409 selected tasks from eight randomly selected papers under both Raw and RRA conditions, totaling 818 outcomes. Review includes outcomes not flagged by either Auditor. Each reviewer traces the measurement to primary execution evidence, records an initial judgment, and cross-checks the other's assessment before they agree on the final label. Selection is at the paper level, and task outcomes can share execution evidence. Appendix~\ref{app:human_audit_protocol} gives the complete protocol.

\begin{figure}[!htb]
\centering
\includegraphics[width=0.75\linewidth]{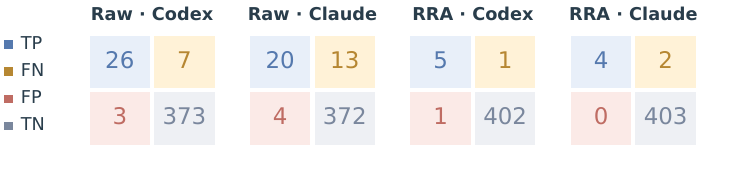}
\caption{\textbf{Audit judgments against human verification.} Two humans cross-verify 409 tasks from eight papers under both Raw and refined-artifact (RRA) conditions (818 outcomes). Each 409-outcome matrix has human invalid/valid rows and Auditor flag/pass columns. TP and FN are detected and missed invalid outcomes; FP and TN are flagged and accepted valid outcomes. Both Auditors are scored against \mbox{the same human labels.}\looseness=-1}
\label{fig:audit_validation}
\end{figure}

Against the shared human labels (Figure~\ref{fig:audit_validation}), Codex achieves 88.6\% precision and 79.5\% recall, versus Claude's 85.7\% and 61.5\%. Each Auditor falsely flags four of 779 valid outcomes (0.5\%). Codex detects seven more invalid outcomes at the same false-positive count. The shared labels identify 39 invalid outcomes: 33 with Raw and six with RRA. Codex misses eight of these outcomes, and Claude misses 15.

In a WBMM timing case, reports replace measured epoch durations with full-dataset projections, including 32.5 seconds reported as 206 seconds; files are edited after process exit. These projections do not constitute the requested measurements. The measured durations also satisfy the numerical intervals, making execution evidence decisive for detecting the substitution. Before invalidation, the Refuter checks the measurement effect and whether permitted protocol choices explain the apparent violation. Appendix~\ref{app:evaluation} presents the human-review protocol and additional audit cases.\looseness=-1

\subsection{Repeated reuse can offset construction API cost}
\label{sec:reuse}
Constructing the 36 artifacts costs \$5,026 in API usage (\$139.61 per paper). At the mean downstream costs in Table~\ref{tab:main}, break-even requires \textbf{8, 13, and 9 reuse rounds} for Sonnet~4.6, Sonnet~5, and Opus~5, respectively. Each round repeats all 36 papers; the scenario charges the full construction cost to one model, holds prices and savings fixed, and excludes GPU, auditing, and maintenance costs. The corresponding API savings are \$653, \$409, and \$615 per complete reuse round, respectively. Higher success accompanies these savings, so reuse improves reproduction while progressively recovering the construction investment. Appendix~\ref{app:cost} details the accounting rates and break-even calculation.

\FloatBarrier
\section{Related work}
\paragraph{Scientific trust and computational reproducibility.}
Scientific trust depends on collective scrutiny and accessible evidence \citep{oreskes2019trust,nosek2015open,munafo2017manifesto}; AI-assisted research retains these requirements \citep{messeri2024illusions}. Following ACM terminology, \emph{reproducibility} means independent execution of authors' artifacts, rather than independent reimplementation or bitwise-identical replay \citep{acm2020artifacts,peng2011reproducible}. Reproducibility Enhancement Principles and the NeurIPS reproducibility program promote explicit dependencies, code release, and reporting checklists \citep{stodden2016enhancing,pineau2021reproducibility}.

\paragraph{Research reproduction benchmarks.}
PaperBench assesses reconstruction through expert rubrics covering code, execution, and results \citep{starace2025paperbench}; SciReplicate-Bench also studies implementation reconstruction \citep{xiang2025scireplicate}. REPRO-Bench and ReplicationBench extend assessment to social science and astrophysics \citep{hu2025reprobench,ye2025replicationbench}. CORE-Bench tests execution of released Code Ocean capsules and questions about their outputs \citep{siegel2024corebench}. \bench{} evaluates released artifacts through binary task success, combining intervals based on variability across five human-verified executions with an independent integrity audit. Tasks include GPU training and evaluation. Table~\ref{tab:corebench} compares handoffs under CORE-Bench's scoring rule.

\paragraph{Automated reproduction and reusable experience.}
PaperCoder constructs repositories through planning and code generation \citep{seo2026paper2code}. AutoReproduce recovers implicit details from cited papers and evaluates execution fidelity \citep{zhao2026autoreproduce}. VERITAS executes research procedures and produces patched code, fix records, and claim assessments \citep{liu2026veritas}; our external comparison tests reuse of its handoff. Reflexion retains verbal feedback, while Voyager accumulates executable skills \citep{shinn2023reflexion,wang2023voyager}. The Compiler for Agent-Native Research Artifacts (ARA) \citep{liu2026ara} packages papers and repositories with expert rubrics in its PaperBench study; a model judge scores reproduction against those rubrics. We construct runnable four-component handoffs without evaluation rubrics or human reference scores, and assess their value through four component removals and an ordinary-outputs control.\looseness=-1

\FloatBarrier
\section{Discussion and limitations}
\label{sec:discussion}
The four component contracts make reproduction experience reusable, and \bench{} provides verified reference scores for evaluation. We study their roles together; the delivery protocol guides what \system{} recovers and preserves during construction. In the formulation of Section~\ref{sec:error_model}, the aim is to reduce avoidable deviations $D$ while documenting uncertainty about the intended experiment.

\subsection{Use before submission and by later reproducers}
The primary intended use is for authors to apply \system{} as a reproducibility check before submitting a paper and releasing its artifacts. Executing the paper's specified setup can expose missing details, execution barriers, and paper--code inconsistencies, giving authors concrete issues to inspect and correct before submission. Third-party reproducers can likewise preserve their work in the same format to lower the cost of later attempts. A human reader can use the setup route and launchers to execute experiments, then inspect the guide and change record to assess the settings and repairs. Our experiments measure reuse by coding agents.

\subsection{Paper--code misalignment}
When the paper and released code conflict, \system{} strictly follows the paper's setup and flags the discrepancy with source locations and affected tasks. For example, the Manifold-Aware Perturbations guide identifies a learning rate of $10^{-4}$ in the paper versus $10^{-3}$ in the released code. Such records help authors check consistency between their manuscript and implementation during the pre-submission review described above.

A separate agent-assisted static analysis records 238 in-scope contradiction findings touching 669 distinct selected tasks. A contradiction requires an explicit paper statement and conflicting implemented behavior; absent implementations and choices on which the paper is silent are recorded separately. These findings are possible sources of execution deviation $D$, but their effects on measured performance require controlled comparisons. They are neither human-validated error rates nor explanations of the measured gains. The analysis is separate from primary success scoring, and its records were not supplied to evaluated Reproducers. Appendix~\ref{app:alignment} documents the scope, categories, and evidence.

\subsection{Limitations}
\label{sec:limitations}
The evaluation conditions on retained papers, fixed reference regions, and the tested agents; it is not a conference-wide reproduction rate. Five reference observations are a protocol choice, not a guarantee of adequate uncertainty estimation. Numerical compatibility does not certify scientific validity or complete fidelity to the stated experiment, and refined artifacts can preserve incorrect assumptions. Automated audits also miss some violations, as quantified against human verification.\looseness=-1

The complete artifact intervention includes upstream investigation, repair, review, and packaging. Ordinary outputs and refined artifacts are produced in separate runs with comparable upstream resource use. This comparison evaluates the construction workflow and resulting handoff together; it does not isolate the effect of reorganizing identical outputs. Component ablations measure losses conditional on the remaining materials. Case annotations and variability strata describe patterns rather than isolate causal effects.

API savings concern downstream Reproducer sessions. Construction, independent verification, GPU execution, maintenance, and human review have separate costs; reuse estimates assume fixed prices and savings. Human time savings and comprehension require a separate study.

\section{Conclusion}
\system{} standardizes recovered reproduction procedures and decisions into four reusable components: a refined codebase, environment setup, experiment launchers, and a reproduction guide. \bench{} provides verified reference scores to assess their value. Reproduction success depends jointly on artifact quality and reproducer ability. All four evaluated models achieve mean reproduction success above 90\% with refined artifacts, including those with lower raw-code success. The cross-model success-rate spread and downstream API spending both decrease. Removing code or environment setup causes the largest ablation losses; removing guides or launchers also reduces success with the other components retained.

Future work should evaluate reuse by human researchers and measure the full costs of constructing, verifying, and maintaining artifacts across repeated attempts.

\section*{AI use statement}
Coding agents constructed the refined artifacts, performed downstream reproductions, and produced automated integrity audits as part of the experiments (Sections~\ref{sec:method} and~\ref{sec:audit}). They also assisted human reference reproducers with setup, debugging, and script preparation under human verification (Appendix~\ref{app:human_reference_protocol}). Agent-assisted reviews annotated recovered outcomes and paper--code discrepancies; these annotations are descriptive and do not establish causal effects or human-validated error rates (Appendices~\ref{app:attribution} and~\ref{app:alignment}).

We used generative AI to draft and edit text, write and check analysis scripts, and prepare figure designs that were redrawn as editable diagrams. The authors defined the research question, hypothesis, benchmark, workflow, and evaluation protocols; reviewed all AI-assisted work; and checked the reported numbers and statistical plots against the recorded results and frozen analysis inputs. We did not use generative AI to generate synthetic data. The authors take responsibility for the final text, claims, and artifacts.

\begingroup
\urlstyle{same}
\bibliographystyle{abbrvnat}
\bibliography{references,reprobridge_bench_50_papers}

@article{pineau2021reproducibility,
  title={Improving Reproducibility in Machine Learning Research (A Report from the {NeurIPS} 2019 Reproducibility Program)},
  author={Pineau, Joelle and Vincent-Lamarre, Philippe and Sinha, Koustuv and Larivi{\`e}re, Vincent and Beygelzimer, Alina and d'Alch{\'e}-Buc, Florence and Fox, Emily and Larochelle, Hugo},
  journal={Journal of Machine Learning Research},
  volume={22}, number={164}, pages={1--20}, year={2021},
  url={https://jmlr.org/papers/v22/20-303.html}
}

@inproceedings{bouthillier2021variance,
  title={Accounting for Variance in Machine Learning Benchmarks},
  author={Bouthillier, Xavier and Delaunay, Pierre and Bronzi, Mirko and Trofimov, Assya and Nichyporuk, Brennan and Szeto, Justin and Mohammadi Sepahvand, Nazanin and Raff, Edward and Madan, Kanika and Voleti, Vikram and Ebrahimi Kahou, Samira and Michalski, Vincent and Arbel, Tal and Pal, Chris and Varoquaux, Ga{\"e}l and Vincent, Pascal},
  booktitle={Proceedings of Machine Learning and Systems},
  volume={3}, year={2021},
  url={https://proceedings.mlsys.org/paper_files/paper/2021/hash/0184b0cd3cfb185989f858a1d9f5c1eb-Abstract.html}
}

@article{nagarajan2018deterministic,
  title={Deterministic Implementations for Reproducibility in Deep Reinforcement Learning},
  author={Nagarajan, Prabhat and Warnell, Garrett and Stone, Peter},
  journal={arXiv preprint arXiv:1809.05676}, year={2018},
  url={https://arxiv.org/abs/1809.05676}
}

@techreport{taylor1994uncertainty,
  title={Guidelines for Evaluating and Expressing the Uncertainty of {NIST} Measurement Results},
  author={Taylor, Barry N. and Kuyatt, Chris E.},
  institution={National Institute of Standards and Technology},
  type={Technical Note}, number={1297}, year={1994},
  doi={10.6028/NIST.tn.1297},
  url={https://doi.org/10.6028/NIST.tn.1297}
}

@misc{acm2020artifacts,
  title={Artifact Review and Badging},
  author={{Association for Computing Machinery}},
  year={2020},
  note={Version 1.1 terminology; accessed November 4, 2026},
  url={https://www.acm.org/publications/policies/artifact-review-and-badging-current}
}

@inproceedings{starace2025paperbench,
  title={{PaperBench}: Evaluating {AI}'s Ability to Replicate {AI} Research},
  author={Starace, Giulio and Jaffe, Oliver and Sherburn, Dane and Aung, James and Chan, Jun Shern and Maksin, Leon and Dias, Rachel and Mays, Evan and Kinsella, Benjamin and Thompson, Wyatt and Heidecke, Johannes and Glaese, Amelia and Patwardhan, Tejal},
  booktitle={Proceedings of the 42nd International Conference on Machine Learning},
  series={Proceedings of Machine Learning Research}, volume={267}, pages={56843--56873},
  publisher={PMLR}, year={2025},
  url={https://proceedings.mlr.press/v267/starace25a.html}
}

@article{siegel2024corebench,
  title={{CORE-Bench}: Fostering the Credibility of Published Research Through a Computational Reproducibility Agent Benchmark},
  author={Siegel, Zachary S. and Kapoor, Sayash and Nadgir, Nitya and Stroebl, Benedikt and Narayanan, Arvind},
  journal={arXiv preprint arXiv:2409.11363}, year={2024},
  note={Version 2, revised June 22, 2026},
  url={https://arxiv.org/abs/2409.11363v2}
}

@article{liu2026ara,
  title={The Last Human-Written Paper: Agent-Native Research Artifacts},
  author={Liu, Jiachen and Pei, Jiaxin and Huang, Jintao and Si, Chenglei and Qu, Ao and Tang, Xiangru and Lu, Runyu and Chen, Lichang and Bai, Xiaoyan and Zheng, Haizhong and Chen, Carl and Chen, Zhiyang and Ye, Haojie and Fu, Yujuan and He, Zexue and Jin, Zijian and Zhang, Zhenyu and Sun, Shangquan and Harmon, Maestro and Wang, John Dianzhuo and Zeng, Jianqiao and Sun, Jiachen and Wu, Mingyuan and Zhou, Baoyu and You, Chenyu and Lu, Shijian and Qiu, Yiming and Lai, Fan and Yuan, Yuan and Li, Yao and Hong, Junyuan and Zhu, Ruihao and Chen, Beidi and Pentland, Alex and Chen, Ang and Chowdhury, Mosharaf and Zhang, Zechen},
  journal={arXiv preprint arXiv:2604.24658}, year={2026},
  note={Version 3, revised May 19, 2026},
  url={https://arxiv.org/abs/2604.24658v3}
}

@article{liu2026veritas,
  title={{VERITAS}: Towards a General-Purpose Replication Tool for Scientific Research},
  author={Liu, Haokun and Tjiaranata, Filbert Aurelian and Tan, Chenhao},
  journal={arXiv preprint arXiv:2607.02931}, year={2026},
  url={https://arxiv.org/abs/2607.02931}
}

@article{shinn2023reflexion,
  title={Reflexion: Language Agents with Verbal Reinforcement Learning},
  author={Shinn, Noah and Cassano, Federico and Berman, Edward and Gopinath, Ashwin and Narasimhan, Karthik and Yao, Shunyu},
  journal={arXiv preprint arXiv:2303.11366}, year={2023},
  url={https://arxiv.org/abs/2303.11366}
}

@article{wang2023voyager,
  title={Voyager: An Open-Ended Embodied Agent with Large Language Models},
  author={Wang, Guanzhi and Xie, Yuqi and Jiang, Yunfan and Mandlekar, Ajay and Xiao, Chaowei and Zhu, Yuke and Fan, Linxi and Anandkumar, Anima},
  journal={arXiv preprint arXiv:2305.16291}, year={2023},
  url={https://arxiv.org/abs/2305.16291}
}

@misc{nistprediction,
  title={Dataplot Reference Manual: Prediction Limits},
  author={{National Institute of Standards and Technology}},
  year={n.d.},
  howpublished={NIST Statistical Engineering Division},
  note={Accessed November 4, 2026},
  url={https://www.itl.nist.gov/div898/software/dataplot/refman1/auxillar/predlimi.htm}
}

@book{oreskes2019trust,
  title={Why Trust Science?},
  author={Oreskes, Naomi},
  publisher={Princeton University Press}, year={2019},
  isbn={9780691179001},
  url={https://press.princeton.edu/books/hardcover/9780691179001/why-trust-science}
}

@article{nosek2015open,
  title={Promoting an open research culture},
  author={Nosek, B. A. and Alter, G. and Banks, G. C. and Borsboom, D. and Bowman, S. D. and Breckler, S. J. and Buck, S. and Chambers, C. D. and Chin, G. and Christensen, G. and Contestabile, M. and Dafoe, A. and Eich, E. and Freese, J. and Glennerster, R. and Goroff, D. and Green, D. P. and Hesse, B. and Humphreys, M. and Ishiyama, J. and Karlan, D. and Kraut, A. and Lupia, A. and Mabry, P. and Madon, T. A. and Malhotra, N. and Mayo-Wilson, E. and McNutt, M. and Miguel, E. and Paluck, E. Levy and Simonsohn, U. and Soderberg, C. and Spellman, B. A. and Turitto, J. and VandenBos, G. and Vazire, S. and Wagenmakers, E. J. and Wilson, R. and Yarkoni, T.},
  journal={Science}, volume={348}, number={6242}, pages={1422--1425}, year={2015},
  doi={10.1126/science.aab2374},
  url={https://doi.org/10.1126/science.aab2374}
}

@article{munafo2017manifesto,
  title={A manifesto for reproducible science},
  author={Munaf{\`o}, Marcus R. and Nosek, Brian A. and Bishop, Dorothy V. M. and Button, Katherine S. and Chambers, Christopher D. and Percie du Sert, Nathalie and Simonsohn, Uri and Wagenmakers, Eric-Jan and Ware, Jennifer J. and Ioannidis, John P. A.},
  journal={Nature Human Behaviour}, volume={1}, pages={0021}, year={2017},
  doi={10.1038/s41562-016-0021},
  url={https://www.nature.com/articles/s41562-016-0021}
}

@techreport{jcgm2020models,
  author={{Joint Committee for Guides in Metrology}},
  title={Guide to the Expression of Uncertainty in Measurement---Part 6: Developing and Using Measurement Models},
  institution={BIPM},
  number={JCGM GUM-6:2020},
  year={2020},
  url={https://www.bipm.org/documents/20126/2071204/JCGM_GUM_6_2020.pdf}
}

@misc{nistmeaninterval,
  author={{NIST/SEMATECH}},
  title={e-Handbook of Statistical Methods: Confidence Limits for the Mean},
  year={n.d.},
  howpublished={Section 1.3.5.2},
  url={https://www.itl.nist.gov/div898/handbook/eda/section3/eda352.htm},
  note={Accessed November 4, 2026}
}

@article{messeri2024illusions,
  title={Artificial intelligence and illusions of understanding in scientific research},
  author={Messeri, Lisa and Crockett, M. J.},
  journal={Nature}, volume={627}, pages={49--58}, year={2024},
  doi={10.1038/s41586-024-07146-0},
  url={https://www.nature.com/articles/s41586-024-07146-0}
}

@article{peng2011reproducible,
  title={Reproducible Research in Computational Science},
  author={Peng, Roger D.},
  journal={Science}, volume={334}, number={6060}, pages={1226--1227}, year={2011},
  doi={10.1126/science.1213847},
  url={https://doi.org/10.1126/science.1213847}
}

@article{stodden2016enhancing,
  title={Enhancing reproducibility for computational methods},
  author={Stodden, Victoria and McNutt, Marcia and Bailey, David H. and Deelman, Ewa and Gil, Yolanda and Hanson, Brooks and Heroux, Michael A. and Ioannidis, John P. A. and Taufer, Michela},
  journal={Science}, volume={354}, number={6317}, pages={1240--1241}, year={2016},
  doi={10.1126/science.aah6168},
  url={https://doi.org/10.1126/science.aah6168}
}

@article{xiang2025scireplicate,
  title={{SciReplicate-Bench}: Benchmarking {LLMs} in Agent-driven Algorithmic Reproduction from Research Papers},
  author={Xiang, Yanzheng and Yan, Hanqi and Ouyang, Shuyin and Gui, Lin and He, Yulan},
  journal={arXiv preprint arXiv:2504.00255}, year={2025},
  url={https://arxiv.org/abs/2504.00255}
}

@inproceedings{hu2025reprobench,
  title={{REPRO-Bench}: Can Agentic {AI} Systems Assess the Reproducibility of Social Science Research?},
  author={Hu, Chuxuan and Zhang, Liyun and Lim, Yeji and Wadhwani, Aum and Peters, Austin and Kang, Daniel},
  booktitle={Findings of the Association for Computational Linguistics: ACL 2025},
  publisher={Association for Computational Linguistics}, pages={23616--23626}, year={2025},
  doi={10.18653/v1/2025.findings-acl.1210},
  url={https://aclanthology.org/2025.findings-acl.1210/}
}

@article{ye2025replicationbench,
  title={{ReplicationBench}: Can {AI} Agents Replicate Astrophysics Research Papers?},
  author={Ye, Christine and Yuan, Sihan and Cooray, Suchetha and Dillmann, Steven and Roque, Ian L. V. and Baron, Dalya and Frank, Philipp and Martin-Alvarez, Sergio and Koblischke, Nolan and Qu, Frank J. and Yang, Diyi and Wechsler, Risa and Ciuca, Ioana},
  journal={arXiv preprint arXiv:2510.24591}, year={2025},
  url={https://arxiv.org/abs/2510.24591}
}

@inproceedings{seo2026paper2code,
  title={{Paper2Code}: Automating Code Generation from Scientific Papers in Machine Learning},
  author={Seo, Minju and Baek, Jinheon and Lee, Seongyun and Hwang, Sung Ju},
  booktitle={International Conference on Learning Representations}, year={2026},
  url={https://proceedings.iclr.cc/paper_files/paper/2026/hash/41b6674c28a9b93ec8d22a53ca25bc3b-Abstract-Conference.html}
}

@inproceedings{zhao2026autoreproduce,
  title={{AutoReproduce}: Automatic {AI} Experiment Reproduction with Paper Lineage},
  author={Zhao, Xuanle and Sang, Zilin and Li, Yuxuan and Shi, Qi and Zhao, Weilun and Wang, Shuo and Zhang, Duzhen and Han, Xu and Liu, Zhiyuan and Sun, Maosong},
  booktitle={Proceedings of the 64th Annual Meeting of the Association for Computational Linguistics (Volume 1: Long Papers)},
  publisher={Association for Computational Linguistics}, pages={21920--21942}, year={2026},
  doi={10.18653/v1/2026.acl-long.1001},
  url={https://aclanthology.org/2026.acl-long.1001/}
}

@inproceedings{li2026text,
  title = {From Text to Forecasts: Bridging Modality Gap with Temporal Evolution Semantic Space},
  author = {Li, Lehui and Wang, Yuyao and Yan, Jisheng and Zhang, Wei and Deng, Jinliang and Sun, Haoliang and Han, Zhongyi and Gong, Yongshun},
  booktitle = {Proceedings of the 43rd International Conference on Machine Learning (ICML)},
  year = {2026},
  url = {https://proceedings.mlr.press/v306/li26hc.html},
  eprint = {2603.12664},
  archivePrefix = {arXiv},
  note = {ICML~2026~Spotlight}
}

@inproceedings{petitbois2026offline,
  title = {Offline Reinforcement Learning of {High-Quality} Behaviors Under Robust Style Alignment},
  author = {Petitbois, Mathieu and Portelas, R\'{e}my and Lamprier, Sylvain},
  booktitle = {Proceedings of the 43rd International Conference on Machine Learning (ICML)},
  year = {2026},
  url = {https://proceedings.mlr.press/v306/petitbois26a.html},
  eprint = {2601.22823},
  archivePrefix = {arXiv},
  note = {ICML~2026~Spotlight}
}

@inproceedings{ferrere2026exact,
  title = {Exact Functional {ANOVA} Decomposition for Categorical Inputs Models},
  author = {Ferrere, Baptiste and Bousquet, Nicolas and Gamboa, Fabrice and Loubes, Jean-Michel and Mur\'{e}, Joseph},
  booktitle = {Proceedings of the 43rd International Conference on Machine Learning (ICML)},
  year = {2026},
  url = {https://proceedings.mlr.press/v306/ferrere26a.html},
  eprint = {2603.02673},
  archivePrefix = {arXiv},
  note = {ICML~2026~Spotlight}
}

@inproceedings{zhou2026distributional,
  title = {A Distributional View for Visual Mechanistic Interpretability: {KL-Minimal} {Soft-Constraint} Principle},
  author = {Zhou, Guancheng and Luo, Yisi and He, Zhengfu and Jin, Zhenyu and Ge, Xuyang and Shu, Wentao and Meng, Deyu and Qiu, Xipeng},
  booktitle = {Proceedings of the 43rd International Conference on Machine Learning (ICML)},
  year = {2026},
  url = {https://proceedings.mlr.press/v306/zhou26ai.html},
  eprint = {2605.17504},
  archivePrefix = {arXiv},
  note = {ICML~2026~Spotlight}
}

@inproceedings{angell2026estimating,
  title = {Estimating Tail Risks in Language Model Output Distributions},
  author = {Angell, Rico and Singhal, Raghav and Horvitz, Zachary and Yu, Zhou and Ranganath, Rajesh and McKeown, Kathleen and He, He},
  booktitle = {Proceedings of the 43rd International Conference on Machine Learning (ICML)},
  year = {2026},
  url = {https://proceedings.mlr.press/v306/angell26a.html},
  eprint = {2604.22167},
  archivePrefix = {arXiv},
  note = {ICML~2026~Spotlight}
}

@inproceedings{li2026initialization,
  title = {Initialization is Half the Battle: Generating Diverse Images from a Guidance Potential Posterior},
  author = {Li, Xiang and Liu, Dianbo and Kawaguchi, Kenji},
  booktitle = {Proceedings of the 43rd International Conference on Machine Learning (ICML)},
  year = {2026},
  url = {https://proceedings.mlr.press/v306/li26de.html},
  eprint = {2606.02453},
  archivePrefix = {arXiv},
  note = {ICML~2026~Spotlight}
}

@inproceedings{zhang2026helix,
  title = {{HELIX:} Hybrid Encoding with Learnable Identity and Cross-dimensional Synthesis for Time Series Imputation},
  author = {Zhang, Fengming and Du, Wenjie and Zhang, Huan and Yu, Ke and Qu, Shen},
  booktitle = {Proceedings of the 43rd International Conference on Machine Learning (ICML)},
  year = {2026},
  url = {https://proceedings.mlr.press/v306/zhang26ak.html},
  eprint = {2605.02278},
  archivePrefix = {arXiv},
  note = {ICML~2026~Spotlight}
}

@inproceedings{howard2026control,
  title = {Control Consistency Losses for Diffusion Bridges},
  author = {Howard, Samuel and N\"{u}sken, Nikolas and Pidstrigach, Jakiw},
  booktitle = {Proceedings of the 43rd International Conference on Machine Learning (ICML)},
  year = {2026},
  url = {https://proceedings.mlr.press/v306/howard26b.html},
  eprint = {2512.05070},
  archivePrefix = {arXiv},
  note = {ICML~2026~Spotlight}
}

@inproceedings{huber2026chebyshev,
  title = {Chebyshev Policies and the Mountain Car Problem: Reinforcement Learning for {Low-Dimensional} Control Tasks},
  author = {Huber, Stefan and Unger, Hannes and Sch\"{a}fer, Georg and Rehrl, Jakob},
  booktitle = {Proceedings of the 43rd International Conference on Machine Learning (ICML)},
  year = {2026},
  url = {https://proceedings.mlr.press/v306/huber26a.html},
  eprint = {2605.22305},
  archivePrefix = {arXiv},
  note = {ICML~2026~Spotlight}
}

@inproceedings{erata2026learning,
  title = {Learning Randomized Reductions},
  author = {Erata, Ferhat and Paradise, Orr and Typaldos, Thanos and Antonopoulos, Timos and Nguyen, ThanhVu and Goldwasser, Shafi and Piskac, Ruzica},
  booktitle = {Proceedings of the 43rd International Conference on Machine Learning (ICML)},
  year = {2026},
  url = {https://proceedings.mlr.press/v306/erata26a.html},
  eprint = {2412.18134},
  archivePrefix = {arXiv},
  note = {ICML~2026~Spotlight}
}

@inproceedings{li2026adaptive,
  title = {Adaptive Testing for {LLM} Evaluation: A Psychometric Alternative to Static Benchmarks},
  author = {Li, Peiyu and Tang, Xiuxiu and Chen, Si and Cheng, Ying and Metoyer, Ronald and Hua, Ting and Chawla, Nitesh V},
  booktitle = {Proceedings of the 43rd International Conference on Machine Learning (ICML)},
  year = {2026},
  url = {https://proceedings.mlr.press/v306/li26fx.html},
  eprint = {2511.04689},
  archivePrefix = {arXiv},
  note = {ICML~2026~Spotlight}
}

@inproceedings{liu2026balancing,
  title = {Balancing Understanding and Generation in Discrete Diffusion Models},
  author = {Liu, Yue and Zhao, Yuzhong and Xie, Zheyong and Ye, Qixiang and Jiao, Jianbin and Hu, Yao and Cao, Shaosheng and Liu, Yunfan},
  booktitle = {Proceedings of the 43rd International Conference on Machine Learning (ICML)},
  year = {2026},
  url = {https://proceedings.mlr.press/v306/liu26gi.html},
  eprint = {2602.01362},
  archivePrefix = {arXiv},
  note = {ICML~2026~Spotlight}
}

@inproceedings{ortiz2026flashoptim,
  title = {{F}lash{O}ptim: Optimizers for Memory-Efficient Training},
  author = {Gonzalez Ortiz, Jose Javier and Gupta, Abhay and Rinard, Christopher and Blalock, Davis},
  booktitle = {Proceedings of the 43rd International Conference on Machine Learning (ICML)},
  year = {2026},
  url = {https://proceedings.mlr.press/v306/gonzalez-ortiz26a.html},
  eprint = {2602.23349},
  archivePrefix = {arXiv},
  note = {ICML~2026~Spotlight}
}

@inproceedings{kowsher2026lime,
  title = {{LiME:} Lightweight Mixture of Experts for Efficient Multimodal Multi-task Learning},
  author = {Kowsher, Md and Mansoor, Haris and Prottasha, Nusrat Jahan and Garibay, Ozlem and Zhu, Victor and Ji, Zhengping and Chen, Chen},
  booktitle = {Proceedings of the 43rd International Conference on Machine Learning (ICML)},
  year = {2026},
  url = {https://proceedings.mlr.press/v306/kowsher26a.html},
  eprint = {2604.02338},
  archivePrefix = {arXiv},
  note = {ICML~2026~Spotlight}
}

@inproceedings{rosa2026guaranteed,
  title = {Guaranteed Optimal Compositional Explanations for Neurons},
  author = {La Rosa, Biagio and Gilpin, Leilani H.},
  booktitle = {Proceedings of the 43rd International Conference on Machine Learning (ICML)},
  year = {2026},
  url = {https://proceedings.mlr.press/v306/la-rosa26a.html},
  eprint = {2511.20934},
  archivePrefix = {arXiv},
  note = {ICML~2026~Spotlight}
}

@inproceedings{keegan2026manifoldaware,
  title = {{Manifold-Aware} Perturbations for Constrained Generative Modeling},
  author = {Keegan, Katherine and Ruthotto, Lars},
  booktitle = {Proceedings of the 43rd International Conference on Machine Learning (ICML)},
  year = {2026},
  url = {https://proceedings.mlr.press/v306/keegan26a.html},
  eprint = {2601.23151},
  archivePrefix = {arXiv},
  note = {ICML~2026~Spotlight}
}

@inproceedings{xue2026anyorder,
  title = {{Any-Order} {GPT} as Masked Diffusion Model: Decoupling Formulation and Architecture},
  author = {Xue, Shuchen and Xie, Tianyu and Hu, Tianyang and Feng, Zijin and Sun, Jiacheng and Kawaguchi, Kenji and Li, Zhenguo and Ma, Zhi-Ming},
  booktitle = {Proceedings of the 43rd International Conference on Machine Learning (ICML)},
  year = {2026},
  url = {https://proceedings.mlr.press/v306/xue26k.html},
  eprint = {2506.19935},
  archivePrefix = {arXiv},
  note = {ICML~2026~Spotlight}
}

@inproceedings{yuan2026markov,
  title = {Markov Chain Monte Carlo without Evaluating the Target: an Auxiliary Variable Approach},
  author = {Yuan, Wei and Wang, Guanyang},
  booktitle = {Proceedings of the 43rd International Conference on Machine Learning (ICML)},
  year = {2026},
  url = {https://proceedings.mlr.press/v306/yuan26j.html},
  eprint = {2406.05242},
  archivePrefix = {arXiv},
  note = {ICML~2026~Spotlight}
}

@inproceedings{macko2026delta4,
  title = {{MACKO}: Sparse matrix-vector multiplication for low sparsity},
  author = {Macko, Vladim\'{\i}r and Bo\v{z}a, Vladim\'{\i}r},
  booktitle = {Proceedings of the 43rd International Conference on Machine Learning (ICML)},
  year = {2026},
  url = {https://proceedings.mlr.press/v306/macko26a.html},
  eprint = {2511.13061},
  archivePrefix = {arXiv},
  note = {ICML~2026~Spotlight}
}

@inproceedings{che2026efficiently,
  title = {Efficiently Training {Time-to-First-Spike} Spiking Neural Networks from Scratch},
  author = {Che, Kaiwei and Ma, Zhengyu and Huang, Yifan and Xue, Peng and Yuan, Li and Fang, Wei and Tian, Yonghong},
  booktitle = {Proceedings of the 43rd International Conference on Machine Learning (ICML)},
  year = {2026},
  url = {https://proceedings.mlr.press/v306/che26a.html},
  eprint = {2410.23619},
  archivePrefix = {arXiv},
  note = {ICML~2026~Spotlight}
}

@inproceedings{wang2026effective,
  title = {Effective Model Pruning: Measuring the Redundancy of Model Components},
  author = {Wang, Yixuan and Guralnik, Dan P. and Akbari, Saiedeh and Dixon, Warren E.},
  booktitle = {Proceedings of the 43rd International Conference on Machine Learning (ICML)},
  year = {2026},
  url = {https://proceedings.mlr.press/v306/wang26bp.html},
  eprint = {2509.25606},
  archivePrefix = {arXiv},
  note = {ICML~2026~Spotlight}
}

@inproceedings{martens2026ocspace,
  title = {{OC-space:} a Unifying Perspective on Verification of Tree Ensembles},
  author = {Martens, Timo and Devos, Laurens and Cascioli, Lorenzo and Meert, Wannes and Blockeel, Hendrik and Davis, Jesse},
  booktitle = {Proceedings of the 43rd International Conference on Machine Learning (ICML)},
  year = {2026},
  url = {https://proceedings.mlr.press/v306/martens26a.html},
  note = {ICML~2026~Spotlight}
}

@inproceedings{lu2026mindomni,
  title = {{Mind-Omni:} A Unified {Multi-Task} Framework for {Brain-Vision-Language} Modeling via Discrete Diffusion},
  author = {Lu, Yizhuo and Du, Changde and Shi, Qingyu and Chen, Hang and Peng, Jie and Jiang, Liuyun and Zhao, Shuangchen and He, Huiguang},
  booktitle = {Proceedings of the 43rd International Conference on Machine Learning (ICML)},
  year = {2026},
  url = {https://proceedings.mlr.press/v306/lu26d.html},
  eprint = {2605.29591},
  archivePrefix = {arXiv},
  note = {ICML~2026~Spotlight}
}

@inproceedings{li2026kineticenergy,
  title = {A Kinetic Energy Perspective of Flow Matching},
  author = {Li, Ziyun and Hu, Huancheng and Lim, Soon Hoe and Li, Xuyu and Gao, Fei and Diao, Enmao and Ding, Zezhen and Vazirgiannis, Michalis and Bostr\"{o}m, Henrik},
  booktitle = {Proceedings of the 43rd International Conference on Machine Learning (ICML)},
  year = {2026},
  url = {https://proceedings.mlr.press/v306/li26bk.html},
  eprint = {2602.07928},
  archivePrefix = {arXiv},
  note = {ICML~2026~Spotlight}
}

@inproceedings{liu2026last0,
  title = {{LaST$_{0}$:} Latent {Spatio-Temporal} {Chain-of-Thought} for Robotic {Vision-Language-Action} Model},
  author = {Liu, Zhuoyang and Liu, Jiaming and Chen, Hao and Yu, Jiale and Guo, Ziyu and Hou, Chengkai and Gu, Chenyang and Mi, Xiangju and Zhang, Renrui and Wu, Kun and Che, Zhengping and Tang, Jian and Heng, Pheng-Ann and Zhang, Shanghang},
  booktitle = {Proceedings of the 43rd International Conference on Machine Learning (ICML)},
  year = {2026},
  url = {https://proceedings.mlr.press/v306/liu26ch.html},
  eprint = {2601.05248},
  archivePrefix = {arXiv},
  note = {ICML~2026~Spotlight}
}

@inproceedings{santis2026mixture,
  title = {Mixture of Concept Bottleneck Experts},
  author = {De Santis, Francesco and Ciravegna, Gabriele and De Felice, Giovanni and Casanova, Arianna and Giannini, Francesco and Diligenti, Michelangelo and Schneider, Johannes and Giordano, Danilo and Espinosa Zarlenga, Mateo and Barbiero, Pietro},
  booktitle = {Proceedings of the 43rd International Conference on Machine Learning (ICML)},
  year = {2026},
  url = {https://proceedings.mlr.press/v306/de-santis26a.html},
  eprint = {2602.02886},
  archivePrefix = {arXiv},
  note = {ICML~2026~Spotlight}
}

@inproceedings{zabergja2026endtoend,
  title = {{End-to-End} Compression for Tabular Foundation Models},
  author = {Zab\"{e}rgja, Guri and Kamel, Rafiq and Kadra, Arlind and Frey, Christian M. M. and Grabocka, Josif},
  booktitle = {Proceedings of the 43rd International Conference on Machine Learning (ICML)},
  year = {2026},
  url = {https://proceedings.mlr.press/v306/zabergja26a.html},
  eprint = {2602.05649},
  archivePrefix = {arXiv},
  note = {ICML~2026~Spotlight}
}

@inproceedings{qiu2026poetx,
  title = {{POET-X:} Memory-efficient {LLM} Training by Scaling Orthogonal Transformation},
  author = {Qiu, Zeju and Liu, Lixin and Weller, Adrian and Shi, Han and Liu, Weiyang},
  booktitle = {Proceedings of the 43rd International Conference on Machine Learning (ICML)},
  year = {2026},
  url = {https://proceedings.mlr.press/v306/qiu26o.html},
  eprint = {2603.05500},
  archivePrefix = {arXiv},
  note = {ICML~2026~Spotlight}
}

@inproceedings{zeng2026ponderlm2,
  title = {{PonderLM-2:} Pretraining {LLM} with Latent Thoughts in Continuous Space},
  author = {Zeng, Boyi and Li, He and Song, Shixiang and Wang, Yixuan and Wang, Zitong and He, Ziwei and Wang, Xinbing and Lin, Zhouhan},
  booktitle = {Proceedings of the 43rd International Conference on Machine Learning (ICML)},
  year = {2026},
  url = {https://proceedings.mlr.press/v306/zeng26l.html},
  eprint = {2509.23184},
  archivePrefix = {arXiv},
  note = {ICML~2026~Spotlight}
}

@inproceedings{wen2026progressive,
  title = {Progressive Graph Structure Adjustment for Homophily Shift Adaptation},
  author = {Wen, Hongwei and Zhang, Can and He, Haoyu and Hang, Hanyuan and Lei, Minglong},
  booktitle = {Proceedings of the 43rd International Conference on Machine Learning (ICML)},
  year = {2026},
  url = {https://proceedings.mlr.press/v306/wen26j.html},
  note = {ICML~2026~Spotlight}
}

@inproceedings{tusoni2026robust,
  title = {Robust Causal Discovery in {Real-World} Time Series with {Power-Laws}},
  author = {Tusoni, Matteo and Masi, Giuseppe and Coletta, Andrea and Glielmo, Aldo and Arrigoni, Viviana and Bartolini, Novella},
  booktitle = {Proceedings of the 43rd International Conference on Machine Learning (ICML)},
  year = {2026},
  url = {https://proceedings.mlr.press/v306/tusoni26a.html},
  eprint = {2507.12257},
  archivePrefix = {arXiv},
  note = {ICML~2026~Spotlight}
}

@inproceedings{zou2026slaclip,
  title = {{SlaClip:} Gradient Norm Slacks can be Indicator for Adaptive Clipping in {DP-SGD}},
  author = {Zou, Shuyan and Wang, Shaowei and Zhu, Zhanxing and Li, Jin and Dong, Changyu and Sassone, Vladimiro and Wu, Han},
  booktitle = {Proceedings of the 43rd International Conference on Machine Learning (ICML)},
  year = {2026},
  url = {https://proceedings.mlr.press/v306/zou26n.html},
  note = {ICML~2026~Spotlight}
}

@inproceedings{brockschmidt2026survdiff,
  title = {{SurvDiff:} A Diffusion Model for Generating Synthetic Data in Survival Analysis},
  author = {Brockschmidt, Marie and Schr\"{o}der, Maresa and Feuerriegel, Stefan},
  booktitle = {Proceedings of the 43rd International Conference on Machine Learning (ICML)},
  year = {2026},
  url = {https://proceedings.mlr.press/v306/brockschmidt26a.html},
  eprint = {2509.22352},
  archivePrefix = {arXiv},
  note = {ICML~2026~Spotlight}
}

@inproceedings{chi2026unifying,
  title = {Unifying and Optimizing Data Values for Selection via Sequential {Decision-Making}},
  author = {Chi, Hongliang and Wu, Qiong and Zhou, Zhengyi and Light, Jonathan and Dodwell, Emily and Ma, Yao},
  booktitle = {Proceedings of the 43rd International Conference on Machine Learning (ICML)},
  year = {2026},
  url = {https://proceedings.mlr.press/v306/chi26d.html},
  eprint = {2502.04554},
  archivePrefix = {arXiv},
  note = {ICML~2026~Spotlight}
}

@inproceedings{kiyani2026trust,
  title = {When to Trust the Cheap Check: Weak and Strong Verification for Reasoning},
  author = {Kiyani, Shayan and Noorani, Sima and Pappas, George J. and Hassani, Hamed},
  booktitle = {Proceedings of the 43rd International Conference on Machine Learning (ICML)},
  year = {2026},
  url = {https://proceedings.mlr.press/v306/kiyani26a.html},
  eprint = {2602.17633},
  archivePrefix = {arXiv},
  note = {ICML~2026~Spotlight}
}

@inproceedings{song2026wbmm,
  title = {{WBMM:} Windowed Batch Matrix Multiplication for Efficient Large Receptive Field Convolution},
  author = {Song, Wan and Zhou, Wei and Wang, Rui and Yu, Jun and Kurihara, Toru and Xu, Jiajia and Zhan, Shu},
  booktitle = {Proceedings of the 43rd International Conference on Machine Learning (ICML)},
  year = {2026},
  url = {https://proceedings.mlr.press/v306/song26ae.html},
  eprint = {2607.02097},
  archivePrefix = {arXiv},
  note = {ICML~2026~Spotlight}
}

@inproceedings{ding2026disentangling,
  title = {Disentangling Latent Risk Pathways via Bayesian Hypergraph Inference},
  author = {Ding, Shengxian and Gao, Haonan and Liu, Pangpang and Tian, Xinyuan and Zhao, Yize},
  booktitle = {Proceedings of the 43rd International Conference on Machine Learning (ICML)},
  year = {2026},
  url = {https://proceedings.mlr.press/v306/ding26c.html},
  eprint = {2606.07677},
  archivePrefix = {arXiv},
  note = {ICML~2026~Spotlight}
}

@inproceedings{supinski2026redhdphmm,
  title = {{RED-HDP-HMM:} {Observation-Dependent} Durations for Bayesian Nonparametric Sequential Models},
  author = {S{\l}upi\'{n}ski, Miko{\l}aj and Lipi\'{n}ski, Piotr},
  booktitle = {Proceedings of the 43rd International Conference on Machine Learning (ICML)},
  year = {2026},
  url = {https://proceedings.mlr.press/v306/slupinski26a.html},
  note = {ICML~2026~Spotlight}
}

@inproceedings{xu2026information,
  title = {Information Flow Reveals When to Trust Language Models},
  author = {Xu, Rui and Chen, Yi and Chen, Jiujiu and Xie, Sihong},
  booktitle = {Proceedings of the 43rd International Conference on Machine Learning (ICML)},
  year = {2026},
  url = {https://proceedings.mlr.press/v306/xu26c.html},
  note = {ICML~2026~Spotlight}
}

@inproceedings{sun2026multimodal,
  title = {Multimodal Latent Language Modeling with {Next-Token} Diffusion},
  author = {Sun, Yutao and Bao, Hangbo and Wang, Wenhui and Peng, Zhiliang and Dong, Li and Huang, Shaohan and Chang, Yaoyao and Wang, Jianyong and Wei, Furu},
  booktitle = {Proceedings of the 43rd International Conference on Machine Learning (ICML)},
  year = {2026},
  url = {https://proceedings.mlr.press/v306/sun26c.html},
  eprint = {2412.08635},
  archivePrefix = {arXiv},
  note = {ICML~2026~Spotlight}
}

@inproceedings{zhao2026transforming,
  title = {Transforming Weather Data from Pixel to Latent Space},
  author = {Zhao, Sijie and Liu, Feng and Zhang, Xueliang and Chen, Hao and Han, Tao and Gong, Junchao and Tao, Ran and Xiao, Pengfeng and Gu, Xinyu and Bai, Lei},
  booktitle = {Proceedings of the 43rd International Conference on Machine Learning (ICML)},
  year = {2026},
  url = {https://proceedings.mlr.press/v306/zhao26ad.html},
  eprint = {2503.06623},
  archivePrefix = {arXiv},
  note = {ICML~2026~Spotlight}
}

@inproceedings{zhang2026subsecond,
  title = {Towards {Sub-Second} Molecular Docking as a Structural Primitive: A Quantized Consistency Diffusion Framework},
  author = {Zhang, Kexin and Qin, Weichen and Teng, Yue and Yu, Jiale and Ma, Yuanyuan and Lin, Jinyu and Sun, Liping and Zheng, Jie and Yu, Jingyi},
  booktitle = {Proceedings of the 43rd International Conference on Machine Learning (ICML)},
  year = {2026},
  url = {https://proceedings.mlr.press/v306/zhang26ew.html},
  note = {ICML~2026~Spotlight}
}

@inproceedings{xie2026mhc,
  title = {{mHC:} {Manifold-Constrained} {Hyper-Connections}},
  author = {Xie, Zhenda and Wei, Yixuan and Cao, Huanqi and Zhao, Chenggang and Deng, Chengqi and Li, Jiashi and Dai, Damai and Gao, Huazuo and Xu, Mingyu and Yu, Kuai and Zhao, Liang and Zhou, Shangyan and Xu, Zhean and Zhang, Zhengyan and Zeng, Wangding and Hu, Shengding and Wang, Yuqing and Yuan, Jingyang and Wang, Lean and Liang, Wenfeng},
  booktitle = {Proceedings of the 43rd International Conference on Machine Learning (ICML)},
  year = {2026},
  url = {https://proceedings.mlr.press/v306/xie26aa.html},
  eprint = {2512.24880},
  archivePrefix = {arXiv},
  note = {ICML~2026~Spotlight}
}

@inproceedings{paulus2026safety,
  title = {Safety Alignment of {LMs} via Non-cooperative Games},
  author = {Paulus, Anselm and Kulikov, Ilia and Amos, Brandon and Munos, R\'{e}mi and Evtimov, Ivan and Chaudhuri, Kamalika and Zharmagambetov, Arman},
  booktitle = {Proceedings of the 43rd International Conference on Machine Learning (ICML)},
  year = {2026},
  url = {https://proceedings.mlr.press/v306/paulus26b.html},
  eprint = {2512.20806},
  archivePrefix = {arXiv},
  note = {ICML~2026~Spotlight}
}

@inproceedings{xie2026controlled,
  title = {Controlled {LLM} Training on Spectral Sphere},
  author = {Xie, Tian and Luo, Haoming and Tang, Haoyu and Hu, Yiwen and Liu, Jason Klein and Ren, Qingnan and Wang, Yang and Zhao, Wayne Xin and Yan, Rui and Su, Bing and Luo, Chong and Guo, Baining},
  booktitle = {Proceedings of the 43rd International Conference on Machine Learning (ICML)},
  year = {2026},
  url = {https://proceedings.mlr.press/v306/xie26r.html},
  eprint = {2601.08393},
  archivePrefix = {arXiv},
  note = {ICML~2026~Spotlight}
}

@inproceedings{chen2026tetrajetv2,
  title = {{TetraJet-v2:} Accurate {NVFP4} Training for Large Language Models with Oscillation Suppression and Outlier Control},
  author = {Chen, Yuxiang and Liu, Yifan and Xu, Xiaoming and Zhang, Pengle and Beyer, Michael and Rapp, Martin and Zhu, Jun and Chen, Jianfei},
  booktitle = {Proceedings of the 43rd International Conference on Machine Learning (ICML)},
  year = {2026},
  url = {https://proceedings.mlr.press/v306/chen26cl.html},
  eprint = {2510.27527},
  archivePrefix = {arXiv},
  note = {ICML~2026~Spotlight}
}

@inproceedings{agrawal2026principled,
  title = {{B}litz{R}ank: Principled Zero-shot Ranking Agents with Tournament Graphs},
  author = {Agrawal, Sheshansh and Nguyen, Thien Hang and Kiela, Douwe},
  booktitle = {Proceedings of the 43rd International Conference on Machine Learning (ICML)},
  year = {2026},
  url = {https://proceedings.mlr.press/v306/agrawal26d.html},
  eprint = {2602.05448},
  archivePrefix = {arXiv},
  note = {ICML~2026~Spotlight}
}

@inproceedings{hong2026maspob,
  title = {{MASPOB:} {Bandit-Based} Prompt Optimization for {Multi-Agent} Systems with Graph Neural Networks},
  author = {Hong, Zhi and Zhang, Qian and Sun, Jiahang and Shang, Zhiwei and Kong, Mingze and Wang, Xiangyi and Shu, Yao and Dai, Zhongxiang},
  booktitle = {Proceedings of the 43rd International Conference on Machine Learning (ICML)},
  year = {2026},
  url = {https://proceedings.mlr.press/v306/hong26l.html},
  eprint = {2603.02630},
  archivePrefix = {arXiv},
  note = {ICML~2026~Spotlight}
}

@inproceedings{he2026clintutorr1,
  title = {{ClinTutor-R1:} Advancing Scalable and Robust {One-to-Many} Alignment in Clinical Socratic Education},
  author = {He, Zhitao and Yang, Haolin and Qin, Zeyu and Fung, Yi R.},
  booktitle = {Proceedings of the 43rd International Conference on Machine Learning (ICML)},
  year = {2026},
  url = {https://proceedings.mlr.press/v306/he26as.html},
  eprint = {2512.05671},
  archivePrefix = {arXiv},
  note = {ICML~2026~Spotlight}
}

@inproceedings{pfaff2026scenesmith,
  title = {{SceneSmith:} Agentic Generation of {Simulation-Ready} Indoor Scenes},
  author = {Pfaff, Nicholas and Cohn, Thomas and Zakharov, Sergey and Cory, Rick and Tedrake, Russ},
  booktitle = {Proceedings of the 43rd International Conference on Machine Learning (ICML)},
  year = {2026},
  url = {https://proceedings.mlr.press/v306/pfaff26a.html},
  eprint = {2602.09153},
  archivePrefix = {arXiv},
  note = {ICML~2026~Spotlight}
}
\endgroup

\clearpage
\appendix
\numberwithin{equation}{section}
\section{Benchmark and reference protocol}
\label{app:selection}
This appendix covers selection, reference verification, statistical rules, and paper inventories.

\subsection{Task selection and reporting units}
\paragraph{Extraction and reporting units.}
Each retained task has a paper identifier, task identifier, source experiment identifier, paper-table locator, reported score, metric direction, task type, and compute type. The selected set contains 730 GPU tasks and 299 CPU-only tasks; 586 have lower-is-better metrics and 443 higher-is-better metrics. Absolute interval membership is used in both cases: outperforming the target in one direction does not automatically satisfy numerical compatibility. Of the 36 papers, 28 are classified as GPU papers and eight as CPU-only papers.

\paragraph{Sampling and exclusions.}
The supplied sampling manifest records 742 Spotlight papers, a mechanically filtered pool of 241 papers with code and non-purely-theoretical content, a random sample of 50, and 36 retained papers. Its primary exclusion categories are artifact/code availability (six), compute requirements (four), and closed-model dependence (four). The compute criteria concern papers for which more than half of the main experiments require over 100 H100 GPU-hours per experiment or more than one eight-H100 node; the API criterion concerns more than half of the main experiments. These are selection criteria, not measured universal execution budgets. Tables~\ref{tab:inventory} and~\ref{tab:excluded} provide the retained and excluded inventories.

\subsection{Primary research-area assignments}
\label{app:domains}
Figure~\ref{fig:domains} uses eight topic families adapted from the ICML~2026 research topics, further divided into 22 subdomains.\footnote{\url{https://icml.cc/Conferences/2026/CallForPapers}} We merge general machine learning with theory, and machine learning systems with optimization; the other six directions remain separate. Each paper receives one editorial assignment by its primary contribution; these are not its original conference submission labels. Generative models fall under deep learning; interpretability, verification, safety, privacy, and causality fall under trustworthy ML. For cross-cutting work, tabular compression is assigned to systems and optimization, conditioned diffusion bridges to probabilistic methods, and brain modeling and survival-data generation to application-driven ML. Table~\ref{tab:domain_assignments} gives the complete primary-area mapping.

\begin{table}[htbp]
\centering\small
\caption{Paper-level assignments to the eight ICML-adapted topic families in Figure~\ref{fig:domains}. IDs refer to the paper inventory in Table~\ref{tab:inventory}. Each paper contributes one count; no task-level weighting is applied.}
\label{tab:domain_assignments}
\begin{tabularx}{\linewidth}{@{}lYr@{}}
\toprule
\rowcolor{tablehead}
Primary research area & Paper IDs & Papers \\
\midrule
Deep learning & 35, 71, 84, 95, 96, 116, 129, 145 & 8 \\
Trustworthy ML & 7, 20, 24, 90, 124, 135, 169, 187 & 8 \\
General ML \& theory & 1, 40, 66, 149, 223, 230 & 6 \\
ML systems \& optimization & 81, 113, 121, 136, 141, 238 & 6 \\
Reinforcement learning \& robotics & 2, 58, 134 & 3 \\
Probabilistic methods & 50, 112 & 2 \\
Application-driven ML & 126, 198 & 2 \\
Evaluation & 69 & 1 \\
 
\bottomrule
\end{tabularx}
\end{table}

\paragraph{Development and test.}
Five NeurIPS~2026 papers and five ICLR~2026 papers supported development of construction, reproduction, and audit workflows. They follow the same task extraction, reference, and integrity-check conventions as test papers. Their results are therefore comparable under the same scoring definition, but they are not held out from workflow development. Results and costs are reported separately in Table~\ref{tab:main} for the 36-paper test set and Appendix~\ref{app:development_results} for development.\looseness=-1

\subsection{Human reference-verification protocol}
\label{app:human_reference_protocol}
For reference collection, a \emph{human verifier} is a participant responsible for both reproducing an experiment and checking its execution evidence. The following requirements apply to each of the five participants in Section~\ref{sec:references}; this role is distinct from human validation of the downstream Auditors.\looseness=-1

\paragraph{Common starting point and independent execution.}
Each verifier must use a separate server and a fresh checkout of the same pinned raw-repository commit, together with the paper, common task specification, and authorized supporting artifacts. Execution directories and coding-agent sessions must remain separate. Verifiers must not reuse another participant's generated scripts or measurements, constructed refined artifacts, or downstream reproduction traces. Published scores remain visible in the paper but cannot substitute for newly computed measurements.

\paragraph{Experimental fidelity and documented repairs.}
Before aggregating the reference measurements, the five verifiers' executed scripts, configurations, and reporting procedures were cross-verified to confirm that the executions followed the paper's specified setup and the common task definition. Verification must check the task's dataset and split, preprocessing, model configuration, training procedure, and evaluation metric against that setup, including required hardware and numerical precision, and inspect the released implementation for consistency. A training task cannot be replaced by checkpoint evaluation; an official checkpoint is permitted when the task explicitly specifies evaluation-only reproduction. Necessary environment or implementation fixes must retain the intended experiment, with their rationale and differences from the starting commit recorded. Material ambiguities must be documented rather than resolved by selecting whichever interpretation best matches the published score.

\paragraph{Human responsibility for agent-assisted work.}
Coding agents may assist with setup, debugging, and script preparation. The human verifier must inspect the scripts and configuration actually executed, including agent-generated edits, data loading, checkpoint selection, and metric extraction. Verification must connect the reported value to the computation that produced it, rather than rely on an agent's summary, a successful exit code, or numerical proximity alone. A measurement whose execution fidelity cannot be established must not be accepted as a reference score.\looseness=-1

\paragraph{Consistent reporting units and repetitions.}
Each participant must produce one complete observation of the specified reporting statistic per task. For a multi-seed mean, this includes all required seed-level runs and their prescribed aggregation; for best-of-$N$, it includes all $N$ candidates and the specified selection rule. Constituent runs must not be pooled across participants to create the five reference observations. Seed choices, repetition counts, and stopping rules must follow the task protocol; unspecified choices must be documented and fixed before inspecting the resulting scores. Prescribed fixed seeds do not remove the requirement for separate executions.

\paragraph{No score-driven selection or substitution.}
Verifiers must not tune procedures or repeat valid measurements until they approach the paper's score. Diagnostic runs and corrected reruns must be distinguished from reference measurements, with the reason for replacement recorded; numerical disagreement alone is not grounds for discarding a valid execution. Failed or resource-blocked attempts must retain their failure reason, without substituting cached scores, another participant's result, or an imputed value. Numerical compatibility is assessed after execution verification using the common rules in Section~\ref{sec:references}.

\paragraph{Evidence retention and information separation.}
Each submitted observation must identify its task, verifier, execution, starting commit, executed scripts and edits, environment and hardware, configuration and seeds, raw outputs, and score-extraction or aggregation procedure. The verification record must state what was checked and any unresolved deviations. Reference measurements and verification guidance must remain separate from \system{} and downstream Reproducer inputs. These requirements support execution fidelity; the statistical assumptions for combining the five observations are specified separately in Appendix~\ref{app:reference_rules}.

\subsection{Error model and reference rules}
\label{app:reference_rules}
We expand the formulation in Section~\ref{sec:error_model} and give additional details of the compatibility and category rules defined in Section~\ref{sec:references}.

\subsubsection{Execution error and reference uncertainty}
\label{app:error_model}
Let $\pi_j$ specify the intended data, configuration, execution conditions, and reporting statistic for task $j$. Its target is $\mu_j=\mathbb E_{\pi_j}[Y_j]$, the expected result under that procedure. A paper score $p_j$ has realized discrepancy $p_j-\mu_j$; neither its magnitude nor its cause is determined by the published value alone. If the reporting unit is a multi-seed mean or best-of-$N$ statistic, $Y_j$ \mbox{denotes that entire statistic.}

For attempts yielding a computed result, let $Z$ denote the realized execution procedure, including choices reconstructed from the supplied information. Expectations describe repeated attempts for a fixed task, agent, and artifact condition. Suppressing these indices, define
\begin{equation}
 m(Z)=\mathbb E[Y\mid Z],\qquad D=m(Z)-\mu,\qquad \varepsilon=Y-m(Z).
 \label{eq:error_components}
\end{equation}
Then $\mathbb E[\varepsilon\mid Z]=0$ and $\mathbb E[D\varepsilon]=0$. With finite second moments, conditioning on the procedure gives\looseness=-1
\begin{equation}
\begin{aligned}
 \mathbb E[(Y-\mu)^2]
 &=\mathbb E[(m(Z)-\mu)^2]+\mathbb E[\operatorname{Var}(Y\mid Z)]\\
 &=(\mathbb E D)^2+\operatorname{Var}(D)+\mathbb E[\varepsilon^2].
\end{aligned}
 \label{eq:mse_decomposition}
\end{equation}
The first two terms capture persistent mean shifts and variation in those shifts across reconstructed procedures. The last captures variation conditional on the procedure. This distinction is analogous to between-condition and within-condition effects in measurement models \citep{jcgm2020models}. For a faithful fixed procedure, $D=0$; conversely, $D=0$ alone does not prove fidelity, since different procedures can have the same expected score. Nor must the residual variance be equal across raw and refined-artifact conditions. Independent repetitions of an unchanged procedure reduce the variance of its mean, not a persistent shift in its expected value.\looseness=-1

\paragraph{What five reference observations estimate.}
Under a faithful reference model, $X_{ji}$ are independent observations of the same reporting statistic under comparable execution conditions, with mean $\mu_j$ and variance $\sigma^2_{j,\mathrm{ref}}$. Let $\bar X_j$ be their sample mean and $s_j$ the observed sample standard deviation. Thus $\mathbb E[\bar X_j]=\mu_j$ and $\operatorname{Var}(\bar X_j)=\sigma^2_{j,\mathrm{ref}}/5$. Approximate normality motivates the Student-$t$ construction. A confidence interval for $\mu_j$ has half-width $t_{0.975,4}s_j/\sqrt5$ \citep{nistmeaninterval}; a prediction interval for a new observation instead has half-width $t_{0.975,4}s_j\sqrt{1+1/5}$ \citep{nistprediction}. The latter includes both reference-center uncertainty and the new observation's variability. Equation~\ref{eq:interval} additionally imposes the explicit 1\% tolerance floor.\looseness=-1

If a downstream measurement $Y$ is independent of these unbiased reference observations, then
\begin{equation}
 \mathbb E[(Y-\bar X_j)^2]
 =\mathbb E[(Y-\mu_j)^2]+\frac{\sigma^2_{j,\mathrm{ref}}}{5}.
 \label{eq:reference_error}
\end{equation}
Squared distance to the reference mean therefore also contains reference-estimation uncertainty. The paper uses the decomposition to motivate knowledge preservation, not to report empirically identified bias or variance components.

\subsubsection{Additional details of the numerical rules}
Section~\ref{sec:references} defines the compatibility interval (Equation~\ref{eq:interval}) and reference-variation measure (Equation~\ref{eq:variation}). The $t$ multipliers for $q=0.95$ and $q=0.80$ are approximately 2.78 and 1.53; calculations retain the stored precision. The 1\% tolerance floor is active on 197 retained tasks. There is no additional near-zero rule for $V_j$: all 21 retained zero-target tasks belong to the higher-variability group. A compatibility interval can contain the reported score even when the five observations do not bracket it.

\subsection{Retained and excluded paper inventories}
\begin{longtable}{@{}rlrrrrr@{}}
\caption{Retained papers and successful task counts from a single evaluation, raw / RRA, under the primary Codex scoring rule. The four columns are Sonnet 4.6 (S4.6), Sonnet 5 (S5), Opus 5 (O5), and Gemini 3.8 Flash (G3.8). IDs are dataset identifiers, not ranks.}\label{tab:inventory}\\
\toprule
\rowcolor{tablehead}
ID & Paper & Tasks & S4.6 & S5 & O5 & G3.8 \\
\midrule
\endfirsthead
\toprule
\rowcolor{tablehead}
ID & Paper & Tasks & S4.6 & S5 & O5 & G3.8 \\
\midrule
\endhead
\bottomrule
\endfoot
1 & \citep{li2026text} & 12 & 9 / 9 & 9 / 9 & 11 / 9 & 12 / 9 \\
2 & \citep{petitbois2026offline} & 15 & 2 / 15 & 15 / 15 & 15 / 15 & 10 / 15 \\
7 & \citep{ferrere2026exact} & 82 & 75 / 76 & 75 / 81 & 80 / 77 & 62 / 75 \\
20 & \citep{zhou2026distributional} & 12 & 10 / 12 & 2 / 12 & 9 / 12 & 11 / 9 \\
24 & \citep{angell2026estimating} & 3 & 3 / 3 & 3 / 2 & 3 / 3 & 3 / 2 \\
35 & \citep{li2026initialization} & 40 & 9 / 40 & 28 / 40 & 40 / 40 & 16 / 40 \\
40 & \citep{zhang2026helix} & 9 & 6 / 9 & 8 / 9 & 4 / 9 & 5 / 6 \\
50 & \citep{howard2026control} & 18 & 15 / 13 & 14 / 16 & 15 / 16 & 11 / 16 \\
58 & \citep{huber2026chebyshev} & 22 & 20 / 20 & 18 / 21 & 20 / 21 & 22 / 22 \\
66 & \citep{erata2026learning} & 38 & 18 / 35 & 16 / 32 & 24 / 35 & 9 / 33 \\
69 & \citep{li2026adaptive} & 82 & 71 / 82 & 55 / 82 & 76 / 82 & 50 / 75 \\
71 & \citep{liu2026balancing} & 21 & 21 / 21 & 18 / 21 & 21 / 21 & 15 / 21 \\
81 & \citep{ortiz2026flashoptim} & 30 & 8 / 30 & 9 / 30 & 30 / 30 & 0 / 29 \\
84 & \citep{kowsher2026lime} & 19 & 4 / 0 & 8 / 4 & 17 / 16 & 4 / 0 \\
90 & \citep{rosa2026guaranteed} & 21 & 0 / 6 & 8 / 14 & 14 / 20 & 3 / 10 \\
95 & \citep{keegan2026manifoldaware} & 60 & 54 / 54 & 54 / 55 & 55 / 55 & 54 / 54 \\
96 & \citep{xue2026anyorder} & 26 & 22 / 25 & 23 / 23 & 25 / 24 & 18 / 21 \\
112 & \citep{yuan2026markov} & 19 & 19 / 7 & 15 / 0 & 18 / 0 & 12 / 12 \\
113 & \citep{macko2026delta4} & 15 & 15 / 15 & 15 / 15 & 15 / 15 & 12 / 15 \\
116 & \citep{che2026efficiently} & 11 & 6 / 11 & 4 / 11 & 6 / 11 & 7 / 11 \\
121 & \citep{wang2026effective} & 46 & 46 / 46 & 46 / 46 & 46 / 46 & 42 / 46 \\
124 & \citep{martens2026ocspace} & 15 & 5 / 13 & 6 / 14 & 11 / 13 & 10 / 11 \\
126 & \citep{lu2026mindomni} & 53 & 43 / 52 & 42 / 53 & 46 / 53 & 53 / 46 \\
129 & \citep{li2026kineticenergy} & 20 & 15 / 19 & 20 / 19 & 18 / 20 & 20 / 18 \\
134 & \citep{liu2026last0} & 16 & 13 / 16 & 12 / 16 & 13 / 16 & 8 / 16 \\
135 & \citep{santis2026mixture} & 22 & 16 / 22 & 17 / 22 & 22 / 22 & 10 / 22 \\
136 & \citep{zabergja2026endtoend} & 45 & 41 / 44 & 41 / 44 & 42 / 44 & 32 / 44 \\
141 & \citep{qiu2026poetx} & 57 & 49 / 57 & 52 / 57 & 57 / 57 & 52 / 55 \\
145 & \citep{zeng2026ponderlm2} & 38 & 37 / 38 & 38 / 38 & 38 / 38 & 37 / 38 \\
149 & \citep{wen2026progressive} & 15 & 7 / 15 & 12 / 15 & 15 / 15 & 15 / 15 \\
169 & \citep{tusoni2026robust} & 23 & 21 / 23 & 23 / 23 & 23 / 23 & 21 / 23 \\
187 & \citep{zou2026slaclip} & 30 & 9 / 24 & 9 / 30 & 12 / 30 & 9 / 30 \\
198 & \citep{brockschmidt2026survdiff} & 14 & 12 / 14 & 14 / 14 & 11 / 14 & 13 / 14 \\
223 & \citep{chi2026unifying} & 13 & 9 / 13 & 9 / 13 & 13 / 13 & 13 / 13 \\
230 & \citep{kiyani2026trust} & 9 & 9 / 9 & 9 / 9 & 9 / 9 & 0 / 9 \\
238 & \citep{song2026wbmm} & 58 & 23 / 57 & 40 / 57 & 44 / 57 & 25 / 57 \\
 
\end{longtable}

\begin{longtable}{@{}rll@{}}
\caption{Excluded papers in the supplied 50-paper manifest. C1: code/artifact availability; C2: node capacity; C3: per-experiment compute; C4: closed-model API dependence. The first listed code is the primary reason. Selection criteria are described in Appendix~\ref{app:selection}.}\label{tab:excluded}\\
\toprule
\rowcolor{tablehead}
ID & Paper & Criteria \\
\midrule
\endfirsthead
\toprule
\rowcolor{tablehead}
ID & Paper & Criteria \\
\midrule
\endhead
\bottomrule
\endfoot
133 & \citep{ding2026disentangling} & C1 \\
156 & \citep{supinski2026redhdphmm} & C1 \\
44 & \citep{xu2026information} & C1 \\
27 & \citep{sun2026multimodal} & C1 / C3 \\
218 & \citep{zhao2026transforming} & C1 / C3 \\
219 & \citep{zhang2026subsecond} & C1 \\
235 & \citep{xie2026mhc} & C3 / C2 / C1 \\
179 & \citep{paulus2026safety} & C3 / C2 / C4 \\
15 & \citep{xie2026controlled} & C3 \\
201 & \citep{chen2026tetrajetv2} & C3 \\
101 & \citep{agrawal2026principled} & C4 \\
39 & \citep{hong2026maspob} & C4 \\
43 & \citep{he2026clintutorr1} & C4 \\
177 & \citep{pfaff2026scenesmith} & C4 \\
 
\end{longtable}
\FloatBarrier

\section{Artifact design and construction}
\label{app:implementation}
This appendix expands the construction contracts in Section~\ref{sec:construction} and Table~\ref{tab:construction_roles} with full role instructions and the artifact's document structure.

\subsection{Construction-skill specifications}
\label{app:workflow_roles}
\system{} implements artifact construction through role-specific skills, each with instructions, reference protocols, tools, and required outputs. The specifications below expand the construction contracts and coordination described in Section~\ref{sec:construction} and Table~\ref{tab:construction_roles}. Separate downstream Reproducer and Auditor workflows evaluate the artifacts; human reference collection and human audit validation belong to benchmark development and evaluation, respectively.

\paragraph{Intake.}
Read the complete paper, including appendices, before planning. Preserve the selected task list and pinned raw commit. Map each task to its table entry, survey the implementation, check required artifacts, and record missing details. Do not silently substitute an unavailable repository revision or redefine a task to fit the available code.

\paragraph{Planner.}
For each task, specify the measurement, data, procedure, aggregation, and stated repetitions, with references to supporting paper passages or code locations. Group tasks that share an execution. Record the adopted resolution of each ambiguity before the dependent run; prepare an executable plan and launcher. The Planner does not execute experiments. When paper and code conflict, strictly follow the paper's stated setup, flag the conflicting code and affected tasks, and record the required implementation corrections before the dependent run.

\paragraph{Builder.}
Resolve and record dependencies, provide reproducible acquisition of required data and checkpoints, and test the resulting environment. The environment described by the deliverable should be the one used for construction measurements, not an untested reconstruction assembled afterward. Environment work must not silently change the experimental plan.

\paragraph{Construction Reproducer.}
Execute planned experiments in the prepared environment. Link values to actual output files and record commands, resources, and failures. Distinguish diagnostic runs from scored repetitions; justify changes and repetition extensions. Do not silently edit a frozen plan or select favorable repetitions in place of the specified reporting statistic. Every task needs a supported measurement or an explicit, evidenced reason why none was produced.

\paragraph{Distiller.}
Write the guide and change record from stored task specifications, execution evidence, repairs, and choices. Explain the reporting statistic, observed variability, resource requirements, pitfalls, and uncovered scope. Preserve uncertainty where the record does not resolve it. Historical measurements belong in separate result records, not in instructions \mbox{substituting for new measurements.}

\paragraph{Packager.}
Assemble the code, environment, launchers, and documentation actually used. Check portable paths and verify the handoff from a clean shell, including an inexpensive real repetition. Compare the output structure and extraction route with the record. Do not invent replacement values or treat a successful smoke test as completion of a scientific experiment.

\paragraph{Reviewer.}
Check the plan before scored execution, inspect evidence and recorded changes during reproduction, and test the final handoff as an unfamiliar user. Inspect the underlying files and outputs directly instead of accepting another role's summary. Produce localized findings with the responsible role; do not repair and approve the same work in one context.

\paragraph{Alignment extension.}
The Aligner skill extracts paper-side requirements before inspecting their implementation, then records contradictions separately from missing code and unspecified details. It does not execute experiments, edit plans, or infer a measured performance effect from a plausible mechanism. Appendix~\ref{app:alignment} describes the separate analysis.

\begin{figure}[p]
\centering
\includegraphics[width=\linewidth,height=0.78\textheight,keepaspectratio]{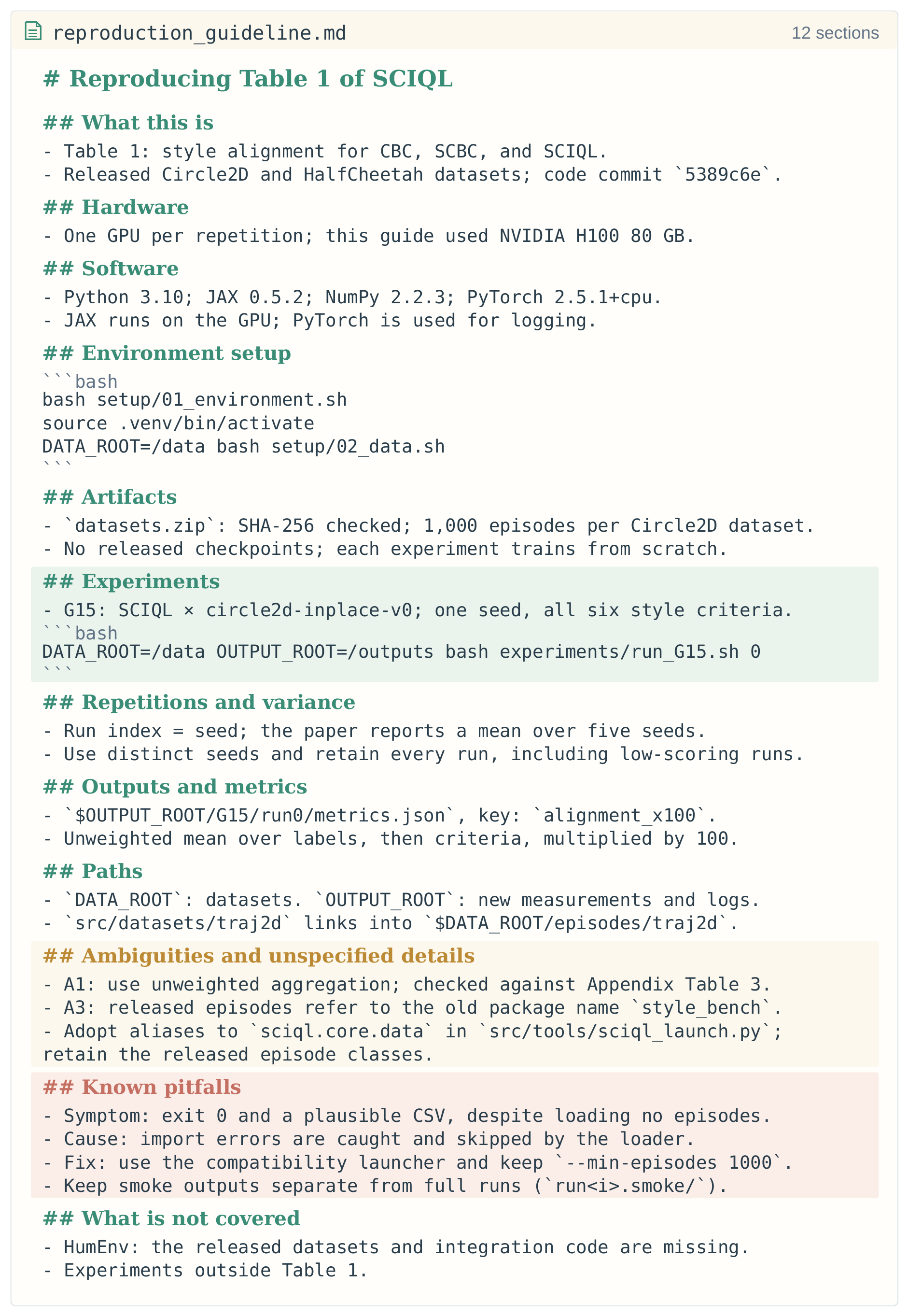}
\caption{\textbf{Inside a reproduction guide.} Condensed SCIQL entries retain all twelve section headings in their original order. The document records hardware and software prerequisites, data acquisition, launch commands, repetitions, output fields and aggregation rules, path conventions, ambiguities, known pitfalls, and unsupported scope. The example links a renamed-package issue to a compatibility launcher and a minimum-episode check, preserving both an executable route and the reason for it. Commands and output keys specify how to produce and extract new measurements.}
\label{fig:guide_appendix}
\end{figure}

\Needspace{12\baselineskip}
\subsection{Four-component handoff and document structure}
\label{app:guide_example}
The normalized refined artifact has seven top-level entries:
\begin{center}
\begin{tabular}{@{}ll@{}}
\toprule
\rowcolor{tablehead}
Path & Role \\
\midrule
\texttt{src/} & Derived source code \\
\texttt{Dockerfile} & Preferred container setup \\
\texttt{setup/} & Environment lock, installation, and data scripts \\
\texttt{experiments/} & Launchers and shared path setup \\
\texttt{reproduction\_guideline.md} & Reproduction guide \\
\texttt{change\_log.md} & Change record and open choices \\
\texttt{verify.sh} & Package verification entry point \\
\bottomrule
\end{tabular}
\end{center}
The archive additionally separates reproduction-result pages and paper--code alignment records. The manuscript's ``change record'' refers to \texttt{change\_log.md}.

The normalized guide has twelve sections: purpose, hardware, software, environment setup, artifacts, experiments, repetitions and variance, expected output structure, paths, ambiguities and unspecified details, pitfalls, and uncovered scope. Its environment section distinguishes Docker from installation scripts. The ambiguity table records ID, type, location, gap, adopted resolution and reason, sensitivity, and affected tasks. The change record uses six sections: changes required to run, changes required for correctness, convenience changes, deliberately unchanged behavior, pitfalls, and open choices.

Across the 36 packaged repositories, normalized documents contain 932 ambiguity-table entries. The corpus contains 482 launcher scripts, including those outside the retained scoring scope.

Figure~\ref{fig:guide_appendix} expands SCIQL's twelve-section guide with representative instructions and decisions.

\section{Evaluation protocol and audit validation}
\label{app:evaluation}
The evaluation uses fresh downstream sessions and a separate integrity audit. The following settings, information boundaries, and validation protocol concern this evaluation, not \mbox{artifact construction.}

Test-set diagnostic statistics here and in Appendix~\ref{app:additional_results} report a single evaluation; Table~\ref{tab:main} and the cost summary in Table~\ref{tab:cost} use means over three evaluations.

\subsection{Model configurations and information boundaries}
\label{app:models}
\begin{table}[htbp]
\centering\small
\caption{Model identifiers and harness settings for main-experiment construction and downstream evaluation. ``Max'' and ``high'' are tool-specific effort settings, not a shared compute scale.}
\begin{tabularx}{\linewidth}{@{}lYll@{}}
\toprule
\rowcolor{tablehead}
Role / label & Model identifier & Harness & Effort \\
\midrule
Construction & \texttt{claude-opus-5[1m]} & Claude Code & max \\
Sonnet 4.6 & \texttt{claude-sonnet-4-6[1m]} & Claude Code & max \\
Sonnet 5 & \texttt{claude-sonnet-5[1m]} & Claude Code & max \\
Opus 5 & \texttt{claude-opus-5[1m]} & Claude Code & max \\
Gemini 3.8 Flash & \texttt{gemini-3.8-flash} & Gemini CLI & high \\
Primary Auditor & \texttt{gpt-5.6-sol} & Codex CLI & high \\
Comparison Auditor & \texttt{claude-opus-5[1m]} & Claude Code & max \\
\bottomrule
\end{tabularx}
\end{table}

\paragraph{Visibility.}
Construction permits the paper, raw repository, required artifacts, and the agents' own execution feedback, while excluding human reference guidance. Reproducers receive the paper and published targets, the selected task specification, and their assigned repository. Reference categories and hidden score fields are removed from session input; the external evaluator retains the exported thresholds. Instructions prohibit using other runs or the alternative repository and require submitted measurements to originate in the current run. Some guides describe historical result directions or magnitudes. The Auditor checks whether submitted measurements come from new executions.

\paragraph{Execution records.}
The downstream Reproducer workflow stores execution IDs, output files and locators, task evidence, and code differences. Where recorded, script/file hashes and ordering checks provide additional provenance evidence. Resource accounting distinguishes wrapped-execution device-hours from total reserved-lane device-hours. Both raw-code and refined-artifact conditions use the same 24-hour wall-clock limit per paper. This session limit applies to downstream evaluation; the experiment-level compute criterion governs benchmark eligibility.

\subsection{Independent reproduction skill}
A fresh downstream Reproducer session uses the \texttt{repro-run} execution skill. It receives the assigned repository, paper, and selected tasks, then plans experiment scripts, prepares permitted inputs, schedules executions, extracts values, and reports code changes and evidence locators. The session retains responsibility for the final evidence, including work delegated by the harness.

Programmatic support freezes inputs, schedules work under the resource limits, links values to files, and checks evidence consistency. Scientific assessment combines these records with inspection of the paper and executed code.

\subsection{Additional audit coordination}
\label{app:audit}
Section~\ref{sec:audit} defines the Finder--Refuter procedure and invalidation criteria; Table~\ref{tab:audit_sensitivity} compares the two Auditors on the same numerically matching outcomes. Codex supplies the primary success labels, while Claude supplies a sensitivity comparison. For either Auditor, success is the number of numerical matches minus invalidations, divided by 1,029 tasks.

Unresolved screening leads require recorded dispositions, and final task-level verdicts are assembled from structured findings. Separate sessions and withheld prior verdicts reduce direct self-validation and cross-auditor anchoring. The human verification protocol below assesses both Auditors against the same evidence requirements.

\subsection{Two-person human verification protocol}
\label{app:human_audit_protocol}
Two human auditors cross-verified selected Reproducer outcomes. The protocol below defines this review's evidence requirements and cross-checking responsibilities. It evaluates existing execution records; human reference collection (Appendix~\ref{app:human_reference_protocol}) \mbox{generates numerical reference observations.}

\paragraph{Review scope and unit.}
Both reviewers must examine each selected task--condition outcome, including outcomes not flagged by either automated Auditor. Reviewing only automated disagreements or invalid flags would miss violations accepted by both systems. Shared executions may support several judgments, but each alleged violation must be linked to the affected task scores; a problem in one experiment does not automatically invalidate every result in the paper.

\paragraph{Trace the measurement to primary evidence.}
Each reviewer must connect the submitted score to the executed command, scripts and configuration, input artifacts, raw outputs, and extraction or aggregation procedure. The supplied code and subsequent changes must be inspected to establish the computation underlying the scored result. Review must also check for copied values, substituted measurements, result-dependent selection, incorrect reductions, and altered execution records. Automated explanations can identify claims to inspect, but are \mbox{not evidence by themselves.}

\paragraph{Apply the same invalidation criteria.}
Reviewers must apply the invalidation criteria in Section~\ref{sec:audit}, considering permitted reporting rules and environment or resource effects before confirming a violation. Judgments concern the executed procedure and its evidence \mbox{rather than inferred intent.}

\paragraph{Cross-check judgments and resolve disagreements.}
Each reviewer must record a judgment and its supporting evidence before the pair compares assessments. Each must then check the other's task mapping, supporting files and outputs, and interpretation, including counterevidence. Disagreements must be resolved by returning to the execution record and documenting why a judgment is retained or revised, not by treating either automated Auditor as a tie-breaker. Both reviewers must endorse the final human label and its rationale; an unresolved case must not be presented as an agreed judgment.

\paragraph{Retain the review record and compare verdicts.}
The record must retain task and condition identifiers, the associated execution, both initial judgments, evidence locators, any disagreement and its resolution, and the final cross-verified label. Correctness of an automated verdict must be evaluated against this human label, separately from whether its explanation identifies the correct mechanism. Count each task--condition outcome once when computing detection statistics, even when several outcomes share execution evidence.

\subsubsection{Human-verified subset}
Eight papers are randomly selected from the test set for two-person cross-verification: ExactFANOVA (82 tasks), Learning Randomized Reductions (38), ATLAS (82), MindOmni (53), POET-X (57), SlaClip (30), Weak-and-Strong Verification (9), and WBMM (58). Review covers all 409 tasks from these papers in both conditions: 409 raw-codebase and 409 refined-artifact outcomes, totaling 818 task--condition outcomes. It identifies 39 invalid outcomes: 33 with raw code and six with refined artifacts. The remaining 779 outcomes are valid under this review, comprising 376 raw-codebase and 403 refined-artifact outcomes. Invalid is the positive class, and uncertain audit judgments are negative. Precision divides correct invalid detections by all invalid predictions; recall divides them by all human-confirmed invalid outcomes. Random selection occurs at the paper level; task outcomes within a paper can share execution evidence and should not be treated as independent draws.

\begin{table}[htbp]
\centering\small
\caption{Audit predictions checked against human verification on 409 outcomes per condition (818 overall). Flagged = TP + FP; TP and FP are correct and incorrect invalid flags, and FN counts missed invalid outcomes. Human-confirmed invalid counts are 33 for Raw and six for RRA. Precision = TP/(TP + FP); recall = TP/(TP + FN).}
\label{tab:audit_validation}
\begin{tabular}{@{}llrrrrrr@{}}
\toprule
\rowcolor{tablehead}
Scope & Auditor & Flagged & TP & FP & FN & Precision (\%) & Recall (\%) \\
\midrule
All & Codex & 35 & 31 & 4 & 8 & 88.6 & 79.5 \\
All & Claude & 28 & 24 & 4 & 15 & 85.7 & 61.5 \\
Raw & Codex & 29 & 26 & 3 & 7 & 89.7 & 78.8 \\
Raw & Claude & 24 & 20 & 4 & 13 & 83.3 & 60.6 \\
RRA & Codex & 6 & 5 & 1 & 1 & 83.3 & 83.3 \\
RRA & Claude & 4 & 4 & 0 & 2 & 100.0 & 66.7 \\
 
\bottomrule
\end{tabular}
\end{table}
Codex misses eight of 39 human-confirmed invalid outcomes (20.5\%); Claude misses 15 (38.5\%). Each incorrectly flags four of 779 human-verified valid outcomes (0.5\%). Codex detects a larger fraction of invalid outcomes in each condition, while Claude makes no false positive on the refined artifact subset (Table~\ref{tab:audit_validation}).

\subsection{Qualitative audit cases}
These cases illustrate measurement validity and evidence integrity through execution records.

\paragraph{Timing reports versus executed measurements: WBMM.}
For refined-artifact tasks T028, T033, and T038, execution records show epoch times of approximately 32.5, 53, and 92 seconds. Submitted JSON and local logs instead contain 206, 335, and 579 seconds, obtained by projecting subset timings to a 2,502-iteration epoch. Reviewed records show file edits 8--17 minutes after process exit and discrepancies with captured stdout. The case review notes that measured durations also satisfy the numerical intervals; the issue is reporting projections as measurements, not missing a target.

\paragraph{Authentic scores do not authenticate the accompanying record.}
In Weak-and-Strong Verification, the case review finds genuine computed scores alongside edited audit records, including a FAIL-to-PASS change and backdated file timestamps. The case demonstrates why a valid numerical computation does not settle whether the submitted evidence faithfully represents the execution.\looseness=-1
\FloatBarrier

\section{Extended results and case evidence}
\label{app:additional_results}
\subsection{Reference-variability strata}
\begin{figure}[htbp]
\centering
\includegraphics[width=0.95\linewidth]{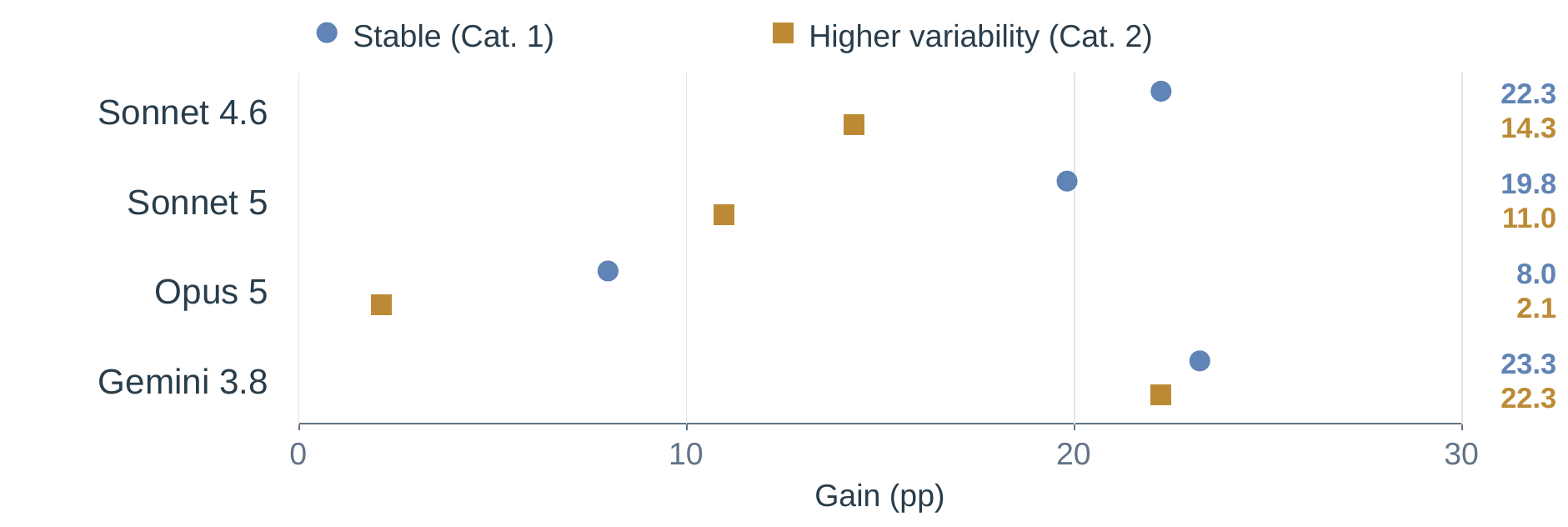}
\caption{\textbf{Gains across reference-variability levels.} Blue circles show stable references (Cat.~1, 701 tasks); amber squares show higher-variability references (Cat.~2, 328 tasks), separated by the 10\% relative-variation threshold. Markers and right-hand labels give refined-artifact (RRA) minus Raw success in percentage points in the first evaluation.}
\label{fig:variability}
\end{figure}
\begin{table}[!htbp]
\centering\small
\caption{\textbf{Success by reference variability in the first evaluation.} Stable (Cat.~1, $n=701$) and higher-variability (Cat.~2, $n=328$) tasks use the same fixed compatibility thresholds in Raw and refined-artifact (RRA) conditions. Gain is RRA minus Raw in percentage points, calculated before rounding the displayed rates.}
\label{tab:strata}
\begin{tabular}{@{}lcrrr@{}}
\toprule
\rowcolor{tablehead}
Model & Cat. & Raw (\%) & RRA (\%) & Gain (pp) \\
\midrule
Sonnet 4.6 & 1 & 71.3 & 93.6 & +22.3 \\
Sonnet 4.6 & 2 & 73.8 & 88.1 & +14.3 \\
Sonnet 5 & 1 & 75.6 & 95.4 & +19.8 \\
Sonnet 5 & 2 & 78.4 & 89.3 & +11.0 \\
Opus 5 & 1 & 88.9 & 96.9 & +8.0 \\
Opus 5 & 2 & 89.9 & 92.1 & +2.1 \\
Gemini 3.8 Flash & 1 & 68.3 & 91.6 & +23.3 \\
Gemini 3.8 Flash & 2 & 66.2 & 88.4 & +22.3 \\
 
\bottomrule
\end{tabular}
\end{table}
Both groups improve for every model, with larger gains in Cat.~1 for each model (Figure~\ref{fig:variability}). Gains span 8.0 to 23.3~pp on stable tasks and 2.1 to 22.3~pp on higher-variability tasks. Wider compatibility intervals can accommodate larger numerical differences, and the task composition differs between strata; the contrast between groups is therefore descriptive.

\Needspace{12\baselineskip}
\subsection{Outcome transitions and execution diagnostics}
\label{app:attribution}
\begin{table}[!htbp]
\centering\small
\setlength{\tabcolsep}{4pt}
\caption{First-evaluation success counts and transitions, with 1,029 tasks per model. Recovery: Raw failure to RRA success; reversal: Raw success to RRA failure. Net is recoveries minus reversals.}
\label{tab:single_evaluation_gains}
\begin{tabular}{@{}lrrrrr@{}}
\toprule
\rowcolor{tablehead}
Model & Raw & RRA & Recoveries & Reversals & Net \\
\midrule
Sonnet 4.6 & 742 & 945 & 234 & 31 & 203 \\
Sonnet 5 & 787 & 962 & 208 & 33 & 175 \\
Opus 5 & 918 & 981 & 100 & 37 & 63 \\
Gemini 3.8 Flash & 696 & 932 & 271 & 35 & 236 \\
\textbf{Total} & \textbf{3,143} & \textbf{3,820} & \textbf{813} & \textbf{136} & \textbf{677} \\
 
\bottomrule
\end{tabular}
\end{table}
\begin{table}[!htbp]
\centering\small
\caption{Downstream repair and time-to-execution diagnostics. Patch counts are papers with a Reproducer code diff out of 36, not independent defects. Time is the median to the first scored execution, \mbox{not end-to-end runtime.}}
\begin{tabular}{@{}lrrrr@{}}
\toprule
\rowcolor{tablehead}
 & \multicolumn{2}{c}{Papers with patches} & \multicolumn{2}{c}{First scored execution (h)} \\
Model & Raw & RRA & Raw & RRA \\
\midrule
Sonnet 4.6 & 26 & 11 & 0.50 & 0.27 \\
Sonnet 5 & 30 & 14 & 0.90 & 0.53 \\
Opus 5 & 24 & 16 & 0.54 & 0.32 \\
Gemini 3.8 Flash & 33 & 13 & 0.73 & 0.26 \\
 
\bottomrule
\end{tabular}
\end{table}

\FloatBarrier
\Needspace{15\baselineskip}
\begin{wraptable}{r}{0.49\textwidth}
\vspace{-\intextsep}
\centering\small
\caption{\textbf{Explanations for recovered outcomes.} Estimated counts allocate case-review shares across 813 recoveries, with rounding that preserves the total.}
\label{tab:attribution}
\begin{tabular}{@{}lrr@{}}
\toprule
\rowcolor{tablehead}
Explanation & Est. outcomes & Share (\%) \\
\midrule
Protocol & 210 & 25.8 \\
Launcher & 76 & 9.4 \\
Env & 87 & 10.7 \\
Code & 99 & 12.2 \\
Artifact & 79 & 9.7 \\
Integrity & 217 & 26.7 \\
Budget & 37 & 4.5 \\
Other & 8 & 1.0 \\
 
\bottomrule
\end{tabular}
\end{wraptable}
An agent-assisted case review assigns primary explanations across 101 model--paper pairs (Table~\ref{tab:attribution}). Analysis agents inspect paired runs and refined-artifact documents; a consolidation agent reopens selected examples, and an independent reviewing agent challenges them. The categories describe explanations grounded in execution records. Their counts allocate the reviewed shares proportionally across recoveries; they are not independent causal measurements.\looseness=-1

\emph{Protocol} covers clarified settings and reporting rules; \emph{Launcher} covers entry points and experiment-to-result mapping. \emph{Env} covers compatible dependencies, and \emph{Code} covers repairs or missing implementations. \emph{Artifact} covers data/checkpoint acquisition and paths. \emph{Integrity} concerns replacing measurement shortcuts with valid execution; audit-invalid names a scored outcome. \emph{Budget} covers resource planning and scheduling; \emph{Other} covers remaining explanations.

\FloatBarrier
\subsection{Failure modes and concentrated regressions}
\label{app:regressions}
\paragraph{Failures shared by both handoffs.}
Qualitative review of task--model pairs that fail with both raw code and refined artifacts identifies paper--code setup mismatches, typographical errors in the paper, and substantial experimental variability. A usable artifact can still preserve a setting that conflicts with the intended experiment or cannot resolve an erroneous paper statement. Even under a fixed procedure, a new observation can fall outside the reference compatibility interval.

\begin{wraptable}{R}{0.51\textwidth}
\vspace{-\intextsep}
\centering\small
\caption{\textbf{Failure states after a reversal.} Counts and shares of all 136 Raw-success-to-RRA-failure outcomes across four models.}
\label{tab:regressions}
\begin{tabular}{@{}lrr@{}}
\toprule
\rowcolor{tablehead}
RRA outcome & Reversals & Share (\%) \\
\midrule
Numerical mismatch & 84 & 61.8 \\
Execution blocked & 10 & 7.4 \\
Resource/access limit & 31 & 22.8 \\
Audit-invalid match & 11 & 8.1 \\
 
\bottomrule
\end{tabular}
\end{wraptable}
\paragraph{Reversals after refinement.}
A reversal is a task--model pair that succeeds with raw code but fails with the refined artifact; it differs from a net decrease over an entire paper. Table~\ref{tab:regressions} partitions all 136 reversals by their refined-artifact outcome. Counts span four models, so a task can appear in several pairs. AuxiliaryMCMC contributes 50, Manifold-Aware Perturbations 19, LiME 14, and ExactFANOVA 10. These four papers account for 93/136 reversals (68.4\%). A paper can contain both recovered and lost tasks: Gemini's AuxiliaryMCMC total remains 12 successes in each condition despite five reversals offset by five recoveries.\looseness=-1

\paragraph{AuxiliaryMCMC: numerical fragility interacts with a finite budget.}
The Opus~5 refined-artifact Gaussian sweep for AuxiliaryMCMC \citep{yuan2026markov} targets 100 rounds at each of three acceptance rates. Its measured workload exceeds the available clock at the chosen concurrency, and the agent executes only 55 rounds per rate. The primary audit invalidates numerical matches affected by this altered measurement procedure. Separately, one Poisson-MALA chain remains constant with zero acceptance. Its undefined effective sample size propagates through the evaluator's ordinary mean, leaving three measurements unavailable. The Reproducer declines to redraw only the failed chain, which would select a repetition based on its outcome. Effective sample size per second also varies with host \mbox{load and concurrency.}

The raw attempt also runs fewer than 100 rounds in part of its sweep. These records illustrate runtime constraints, numerical edge cases, and deviations from the prescribed \mbox{measurement procedure.}

\paragraph{LiME: setup success is not experiment completion.}
In Sonnet~5's refined-artifact attempt on LiME \citep{kowsher2026lime}, data staging and environment preparation take approximately 83 minutes, but the full schedule includes 5,949 training steps followed by evaluation over 47 splits. The attempted run's observed throughput projects 7 to 92 wall-clock hours per variant, competing within the paper's 24-hour allocation. Model-construction parameter counts are available before training finishes, whereas accuracy measurements require completed training and evaluation. The session reaches the former but not the latter. Across models, 13 of LiME's 14 reversals are resource/access failures and one is a numerical mismatch.

\paragraph{Interpreting residual numerical mismatches.}
The 84 numerical reversals span several possible mechanisms: stochastic training, timing conditions, changed execution choices, and unresolved paper--code disagreements. Construction may also introduce an incorrect setting, metric interpretation, or repair that later agents inherit. Diagnosing a particular reversal requires tracing its executed configuration and measurement procedure. The Manifold-Aware Perturbations guide, for example, flags a learning-rate discrepancy between the paper and released code for further inspection (Appendix~\ref{app:examples}).

\subsection{Component ablations and reconstruction effort}
\label{app:ablation}
We randomly select 12 papers from the 36-paper test set for the Sonnet~5 ablation. The same 287 tasks are evaluated across all seven conditions in Table~\ref{tab:ablation}; Table~\ref{tab:ablation_subset} lists the selected papers and their experiment and task counts.

\begin{table}[htbp]
\centering\small
\caption{\textbf{Randomly selected papers for component ablations.} IDs match Table~\ref{tab:inventory}. Tasks are scored measurements associated with the listed experiments; CPU-only tasks are a subset of Tasks. The same 12 papers and 287 tasks are used in every condition.}
\label{tab:ablation_subset}
\begin{tabular}{@{}rlrrr@{}}
\toprule
\rowcolor{tablehead}
ID & Paper & Experiments & Tasks & CPU-only tasks \\
\midrule
1 & TESS & 4 & 12 & 0 \\
20 & EnergyDPS & 3 & 12 & 6 \\
50 & ControlConsistencyBridges & 13 & 18 & 0 \\
112 & AuxiliaryMCMC & 2 & 19 & 19 \\
129 & KPEKTS & 5 & 20 & 0 \\
141 & POET-X & 57 & 57 & 0 \\
149 & PSAHS & 15 & 15 & 0 \\
169 & PLACy & 13 & 23 & 23 \\
187 & SlaClip & 30 & 30 & 0 \\
198 & SurvDiff & 3 & 14 & 0 \\
230 & Weak-and-Strong Verification & 1 & 9 & 9 \\
238 & WBMM & 9 & 58 & 0 \\
 
\midrule
\rowcolor{tablerefined}
 & \textbf{Total} & \textbf{155} & \textbf{287} & \textbf{57} \\
\bottomrule
\end{tabular}
\end{table}

For ordinary outputs, Opus~5 attempts reproduction with the raw codebase. Its execution directory, with result files removed but code, scripts, and logs retained, becomes the input codebase for a fresh Sonnet~5 session; no standardized documentation is added. This run and \system{} construction are separate upstream attempts with comparable resource use. \system{} receives the four-component delivery protocol from the start; the comparison evaluates \mbox{the two workflows' handoffs.}

Section~\ref{sec:ablation} defines the component removals and explains why descriptions of removed components are also deleted. The following case shows the reconstruction work that can return after removing the prepared environment.

\paragraph{WBMM without the prepared environment.}
For WBMM \citep{song2026wbmm}, the environment ablation retains experiment code and commands but removes the compatible H100 dependency route. Sonnet~5 reconstructs a Python~3.10 / PyTorch~2.0 / CUDA~11.8 stack with compatible vision libraries. A supplied binary lacks an operator's H100 kernels, requiring a source rebuild before COCO evaluation. The agent also fixes an installation-tool compatibility issue and moves data off a quota-constrained shared filesystem. No paper-source patch is required, yet substantial environment work returns downstream. The trace produces measurements for \mbox{all 58 selected tasks.}

Timing measurements also depend on GPU concurrency and storage behavior. In this run, reuse of a smaller local image pool changes cache behavior alongside the reconstructed environment.

\subsection{Worked reproduction examples}
\label{app:examples}
These examples expand the paired-run case evidence in Section~\ref{sec:mechanisms}, retaining exact configurations, task identifiers, and measurements. They illustrate information preserved in the refined artifact and corresponding recorded outcomes.

\paragraph{Recovering omitted configuration: HELIX.}
\label{app:helix_case}
For HELIX \citep{zhang2026helix}, Opus~5 restores the paper's architecture on raw code but retains a shipped batch size of 32 and learning rate of approximately $6.29\times10^{-4}$. The guide's ambiguity entry A6 identifies revision \texttt{9760b347}, whose architecture matches the paper configuration, as the source for batch size 8 and learning rate $5.75\times10^{-4}$. The \texttt{--paper\_config} launcher makes this choice executable. The recorded MAE changes from 0.15265 to 0.12986 for T004 and from 0.50527 to 0.41160 for T006; the respective compatibility intervals are [0.12524, 0.13448] and [0.38442, 0.43878]. The artifact records the repository history and architectural match as the rationale for adopting these settings.

\paragraph{An environment route that transfers: WBMM.}
\label{app:wbmm_case}
For six COCO tasks (T063--T068) in WBMM \citep{song2026wbmm}, Sonnet~4.6's raw attempt fails to install a compatible \texttt{mmcv-full} stack for H100. The refined artifact specifies matching PyTorch/MMDetection dependencies, the Ninja build prerequisite, and a source build for the target architecture. The downstream refined-artifact session makes no code patch and executes all six tasks. Its two box AP values, 51.61 and 52.78, agree numerically with reported targets 51.6 and 52.8; mask AP is 44.78/45.64 against 44.8/45.6. The artifact preserves an executable dependency route for later reuse.

\paragraph{Fixing validation labels: Effective Model Pruning.}
The released TinyImageNet loader uses \texttt{ImageFolder} on a validation directory containing only an \texttt{images} subdirectory. It therefore assigns all 10,000 validation images to one class. During construction, \system{} adds a preparation script that places images into class directories using \path{val_annotations.txt}, verifies 200 classes and 10,000 images, and writes a preparation receipt required by the launcher. In Gemini~3.8 Flash's paired runs, the ResNet-18 validation loss is 8.898 with raw code and 2.288 with the refined artifact, against the paper's 2.303. This records a concrete repair and its downstream reuse.

\paragraph{Organizing measurement code for reuse: ETTFS.}
The raw Opus~5 attempt searches the released ETTFS \citep{che2026efficiently} source for the metrics in Tables~2 and~3, finds no producing implementation, and leaves the inference-time measurement unexecuted. During construction, \system{} adds \path{Code/profile_ttfs.py}, documents the metric definitions and output fields, and supplies a launcher that trains a checkpoint before profiling it. The later Opus~5 session uses this explicit route and reports 2.0652 inference time steps against the paper's 2.03. The artifact localizes code and resolved decisions that would otherwise require renewed investigation.

\paragraph{A measurement is not just a model execution: Mixture of Concept Bottleneck Experts.}
For four-expert settings in Mixture of Concept Bottleneck Experts \citep{santis2026mixture}, the paper does not specify how multiple expression distances should be reduced, and the repository provides no Table~1 collector. The raw Sonnet~5 attempt constructs a matching between learned and reference expressions but reports the sum of tree-edit distances. For two tasks this produces 60 and 188, whereas the per-expression means are 15 and 47. The refined artifact resolves the ambiguity by documenting the denominator and implementing the corresponding collector; those downstream values are 15 and 47. The case locates the discrepancy in the underspecified reporting statistic.

\paragraph{Preserving the experimental aggregation: SCIQL.}
The released SCIQL \citep{petitbois2026offline} code produces run outputs but does not assemble the Table~1 statistic, leaving the mapping from logged fields and experiment settings to reported values incomplete. The raw Sonnet~4.6 attempt uses one default criterion, one seed, and an incorrect output column for a table statistic that averages across six criteria and five seeds. The refined artifact's launcher and collector make this mapping explicit by specifying the sweep and the correct alignment field. One representative measurement changes from 19.70 to 55.66 against a paper value of 58.6. Criterion selection, output mapping, and repetition change together in this case, illustrating preservation of the complete measurement protocol.

\paragraph{An optimizer release is not a complete experiment: FlashOptim.}
FlashOptim \citep{ortiz2026flashoptim} supplies optimizer code but not the GPT-2 training and evaluation harness for the selected table. Gemini's raw attempt leaves 18 GPT-2 outcomes unexecuted. The refined artifact adds a training/evaluation harness, data acquisition, output mapping, and launchers; all 18 pass in its recorded downstream attempt. The construction report lists 6,346 added \mbox{lines of experiment support.}

\paragraph{A localized repair: POET-X.}
The defect occurs in \path{examples/peft_pretraining/training_utils.py}, line~222. Full batches call \texttt{collate\_fn}, whereas the trailing partial batch is returned as a list. The evaluator calls \texttt{batch.items()}, so it fails when the final batch is reached. The patch changes the tail to \texttt{yield collate\_fn(batch)}. The record traces the sole caller to validation and describes CPU checks covering empty, partial, complete, and over-full batch counts, without changing the example count. This example illustrates how a later reproducer can recover both the fix and its experimental scope.

\paragraph{Flagging a setup discrepancy: Manifold-Aware Perturbations.}
The guide's ambiguity entry A1 contrasts the paper's stated learning rate of $10^{-4}$ with the released driver, training scripts, and 171 committed manifests using $10^{-3}$, and marks the discrepancy as high sensitivity. This gives authors a concrete inconsistency to inspect and resolve before submission.

\subsection{Separate alignment analysis}
\label{app:alignment}
A separate static audit compares selected paper results with their raw implementations. It distinguishes implemented behavior that contradicts the paper from absent implementations and details on which the paper is silent. Among 293 recorded findings, 238 are in scope for contradiction analysis and touch 669 distinct tasks. High-severity labels occur on 73 findings touching 482 tasks; 25 papers have at least one such finding. Task counts are deduplicated; a task may have multiple severity labels.

Agents produced these findings through static inspection of papers, code, and committed evidence. The analysis is separate from primary success scoring and covers the selected evaluation tasks. The evidence includes the learning-rate case above and a diffusion-objective case in which the code squares a norm described as an unsquared distance, changing its gradient. These cases illustrate that numerical compatibility and code execution can coexist with \mbox{a paper--implementation discrepancy.}

\paragraph{From an apparent gap to a localized finding.}
The audit first extracts the paper-side requirement independently of the code-side inspection. It then identifies the implementation in that role, such as the function computing the loss or selecting the evaluation population. A contradiction requires an explicit paper statement and conflicting implemented behavior, with source locations and affected task IDs. If no implementation occupies the role, the finding is a missing implementation; if the paper specifies no relevant choice, it is underspecification rather than a contradiction. This separates 44 absent-implementation records and 11 paper-silent records from the 238 in-scope findings.

\begin{table}[htbp]
\centering\small
\caption{Static paper--code contradiction findings by recorded category. Scope is determined from evidence, not category names. Distinct tasks and papers are deduplicated within a row but can overlap between rows; these columns must not be summed. Counts describe agent-reviewed records.}
\label{tab:alignment_categories}
\begin{tabular}{@{}lrrr@{}}
\toprule
\rowcolor{tablehead}
Category & Findings & Papers & Distinct tasks \\
\midrule
Hyperparameters & 41 & 26 & 287 \\
Objective & 41 & 22 & 400 \\
Data & 40 & 20 & 298 \\
Architecture & 33 & 18 & 209 \\
Evaluation procedure & 30 & 20 & 261 \\
Training procedure & 22 & 16 & 176 \\
Metric & 21 & 13 & 112 \\
Baseline implementation & 5 & 3 & 4 \\
Divergent implementation choices & 3 & 3 & 15 \\
Scale / hardware & 2 & 2 & 25 \\
 
\bottomrule
\end{tabular}
\end{table}

\paragraph{What these discrepancies mean.}
A learning-rate disagreement is concrete even without retraining: one can point to the paper's $10^{-4}$ and the code's $10^{-3}$. But the resulting change in final accuracy cannot be inferred from that factor alone. Similarly, replacing a distance by its square changes the objective and gradient weighting; its empirical impact requires a controlled comparison. The workflow records the contradiction and its scope while following the paper's specified setup. These records can guide authors' reproducibility checks before submission. In the terms of Section~\ref{sec:error_model}, the discrepancies are possible sources of $D$, not estimates of its magnitude.

Severity is also an agent judgment. The record set assigns 73 findings a high label, of which 63 use an argued magnitude-based route and 10 concern inability to instantiate the stated setup. These labels help prioritize inspection. The alignment records were not supplied to \mbox{the evaluated Reproducers.}

\subsection{Development-set results}
\label{app:development_results}
We use Sonnet~5 and Gemini~3.8 Flash as Reproducers on the ten development papers: five from NeurIPS~2026 and five from ICLR~2026. Table~\ref{tab:development} reports a single evaluation of reproduction success and downstream API spending under the same scoring definition as the test set. These papers informed workflow development, and their results are kept separate from the test-set aggregates.\looseness=-1

\begin{table}[!htbp]
\centering\small
\caption{\textbf{Development-set results from a single evaluation.} Success and downstream Reproducer API spending on the ten papers used for workflow development, reported separately from test results. RRA denotes refined reproducible artifacts; gain is RRA minus Raw in percentage points, and API reduction is $100(\mathrm{Raw}-\mathrm{RRA})/\mathrm{Raw}$. Construction, auditing, and GPU costs are excluded.}
\label{tab:development}
\begin{tabular}{@{}lrrrrrr@{}}
\toprule
\rowcolor{tablehead}
 & \multicolumn{3}{c}{Success rate (\%)} & \multicolumn{3}{c}{API USD} \\
\cmidrule(lr){2-4}\cmidrule(l){5-7}
Reproducer & Raw & RRA & Gain (pp) & Raw & RRA & Reduction (\%) \\
\midrule
Sonnet 5 & 75.6 & \textbf{92.2} & +16.6 & 1,883 & 1,396 & 25.9 \\
Gemini 3.8 Flash & 59.7 & \textbf{91.2} & +31.5 & 478 & 389 & 18.6 \\
 
\bottomrule
\end{tabular}
\end{table}
\FloatBarrier

\section{Cost accounting}
\label{app:cost}
\begin{table}[htbp]
\centering\small
\setlength{\tabcolsep}{4pt}
\caption{Three-evaluation mean paired Reproducer costs from Table~\ref{tab:main}, raw / RRA. Each model uses the same papers in both arms. Tokens include recorded input, output, cache writes/reads, and separately recorded thinking; USD uses frozen accounting rates. Construction, audit, and GPU costs are excluded. Reductions use the means.}
\label{tab:cost}
\begin{tabular}{@{}lrrrr@{}}
\toprule
\rowcolor{tablehead}
Model & Tokens (B) & Reduction (\%) & API USD & Reduction (\%) \\
\midrule
Sonnet 4.6 & 8.48 / 6.80 & 19.8 & 3,165 / 2,512 & 20.6 \\
Sonnet 5 & 4.98 / 4.52 & 9.2 & 1,726 / 1,317 & 23.7 \\
Opus 5 & 3.01 / 2.82 & 6.3 & 3,480 / 2,865 & 17.7 \\
Gemini 3.8 Flash & 2.24 / 1.83 & 18.3 & 392 / 326 & 16.8 \\
 
\bottomrule
\end{tabular}
\end{table}

Table~\ref{tab:cost} reports three-evaluation mean Reproducer costs, summing paper sessions within each evaluation. Claude tokens comprise input, output (including thinking), cache writes, and cache reads. Gemini totals include its separately recorded thinking field. The fixed per-million-token accounting rates, ordered as input/output/cache-write/cache-read, are $(3,15,3.75,0.30)$ for Sonnet~4.6, $(2,10,2.50,0.20)$ for Sonnet~5, and $(5,25,6.25,0.50)$ for Opus~5. Gemini rates are \$0.75 for input, \$3.75 for output plus thinking, and \$0.075 for cache reads; \mbox{recorded cache writes are zero.}

Paper-session costs cover failed attempts and work beyond the 1,029 retained scoring tasks.

\subsection{Construction investment and reuse}
\label{app:reuse}
Constructing the 36 artifacts costs \$5,026 in API usage, averaging \$139.61 per paper.

\paragraph{Sources of construction effort.}
The workflow provides a qualitative explanation of API expenditure. Specification requires reading the paper and code to resolve experiment settings and measurement rules. Execution support requires environment repair, missing implementation, and diagnosis through successive runs; each revision can require further code and log inspection. Review and packaging then check the evidence and preserve the resulting procedures. The WBMM dependency route and HELIX configuration recovery in Section~\ref{sec:mechanisms}, together with the FlashOptim training/evaluation harness in Appendix~\ref{app:examples}, illustrate these kinds of work. GPU runtime is a separate expense: it affects API spending through associated model calls for setup, inspection, and debugging, not through elapsed GPU time itself.

\paragraph{Model-specific reuse.}
For downstream model $m$ in \{Sonnet~4.6, Sonnet~5, Opus~5\}, let $c_m^{\mathrm{raw}}$ and $c_m^{\mathrm{refined}}$ be the mean paired aggregate API costs in Table~\ref{tab:main}, and let $C_{\mathrm{build}}=\$5{,}026$ be the full construction expense. One reuse round repeats the 36-paper workload represented by those costs. With stationary per-round spending, raw reuse costs $n c_m^{\mathrm{raw}}$, while refined-artifact reuse costs $C_{\mathrm{build}}+n c_m^{\mathrm{refined}}$. The API break-even count is
\[
 n_m^*=\left\lceil \frac{C_{\mathrm{build}}}{c_m^{\mathrm{raw}}-c_m^{\mathrm{refined}}}\right\rceil
 \quad\text{when }c_m^{\mathrm{raw}}>c_m^{\mathrm{refined}}.
\]
Each model-specific scenario bears the full construction cost. For positive construction cost and nonpositive savings, there is no finite API crossover.

For example, Sonnet~5 costs a mean of \$1,726 per raw round and \$1,317 per refined-artifact round. The difference is \$409; \$5,026 divided by this saving is 12.29 rounds, hence 13 complete rounds to break even. The corresponding continuous crossover points are 7.70 rounds for Sonnet~4.6 and 8.17 for Opus~5, giving the 8 and 9 complete rounds reported in Section~\ref{sec:reuse}.

These scenarios repeat the complete 36-paper workload at fixed prices, caching, artifact contents, and agent behavior. Costs and savings are aggregated over paper sessions, including shared executions that produce multiple task scores. The crossover accounts for construction and downstream Reproducer API spending; independent auditing, GPU execution, maintenance, and unrecorded cache-storage fees are excluded. Human reference collection is a separate benchmark-development cost.
\FloatBarrier

\end{document}